\documentclass[twoside]{article}

\usepackage{fullpage}

\newcommand{\figdir}{figures}

\newcommand{\myeat}[1]{}

\usepackage[usenames,dvipsnames]{xcolor}
\usepackage{longtable}

\usepackage[toc,page]{appendix}

\usepackage{bm}
\usepackage{frcursive} 
\usepackage{xfrac}

\usepackage{fancyvrb} 
\usepackage{xspace}
\usepackage{amsmath}
\usepackage{amssymb}
\usepackage{amsthm}

\usepackage[mathscr]{euscript}

\mathchardef\mhyphen="2D 
\mathchardef\mdash="2D 

\usepackage{accents}

\usepackage[font={small,it}]{caption}
\usepackage{subcaption} 

\usepackage{url}
\usepackage{cancel} 
\usepackage{multirow}
\usepackage{multicol}

\usepackage{verbatim}

\usepackage[T1]{fontenc}    
\usepackage{url}            
\usepackage{booktabs}       
\usepackage{amsfonts}       
\usepackage{nicefrac}       
\usepackage{microtype}      

\usepackage{booktabs}
\usepackage{comment}

\usepackage{paralist} 

\usepackage{tikz}
\usetikzlibrary{fit,arrows,calc,positioning,shadows.blur}

\usepackage{mathtools}

\usepackage{enumitem}

\newcommand{\dontprintsemicolon}{}
\usepackage[ruled,vlined,linesnumbered]{algorithm2e}

\usepackage{mdframed}
\usepackage{multicol}
\usepackage{rotating}
\usepackage{setspace}

\usepackage{blkarray}

\usepackage[utf8]{inputenc}
\DeclareUnicodeCharacter{00B4}{'}

\usepackage{imakeidx}
\usepackage[columns=3]{idxlayout}

\usepackage{hyperref}

\usepackage[capitalize]{cleveref}
\Crefname{equation}{Equation}{Equations}
\crefname{equation}{Equation}{Equations}
\Crefname{section}{Section}{Sections}
\crefname{section}{Section}{Sections}
\Crefname{chapter}{Chapter}{Chapters}
\crefname{chapter}{Chapter}{Chapters}
\Crefname{figure}{Figure}{Figures}
\crefname{figure}{Figure}{Figures}
\Crefname{appsec}{Appendix}{Appendices}
\crefname{appsec}{Appendix}{Appendices}

\definecolor{cyan}{cmyk}{1,0,0,0.4}
\definecolor{magenta}{cmyk}{0,1,0,0.15}
\definecolor{yellow}{cmyk}{0,0,1,0}
\definecolor{black}{cmyk}{0,0,0,1}
\definecolor{orange}{cmyk}{0,0.5,1,0}
\definecolor{red}{cmyk}{0.00,0.92,0.92,0.06}
\definecolor{blue}{cmyk}{0.92,0.92,0.00,0.06}
\definecolor{green}{cmyk}{0.88,0.00,0.78,0.03}

\hypersetup{
    colorlinks,
    citecolor=magenta,
    linkcolor=magenta,
    urlcolor=cyan,
}

\usepackage{eqnalign}
\usepackage{version}

\newcommand{\freward}{f^{\text{dnn}}}

\newcommand*{\fullref}[1]{\hyperref[{#1}]{\cref{#1} (\nameref*{#1})}}

\newcommand{\eat}[1]{}

\mathchardef\mhyphen="2D 
\mathchardef\mdash="2D 

\newcommand{\myvec}[1]{\mathbf{#1}}
\newcommand{\myvecsym}[1]{\boldsymbol{#1}}

\makeatletter
\newcommand{\oset}[3][-0.3ex]{%
  \mathrel{\mathop{#3}\limits^{
    \vbox to#1{\kern-4\ex@
    \hbox{$\scriptstyle#2$}\vss}}}}
\makeatother

\newcommand{\vzero}{\myvecsym{0}}

\newcommand{\vmu}{\myvecsym{\mu}}

\newcommand{\vLambda}{\myvecsym{\Lambda}}

\newcommand{\vphi}{\myvecsym{\phi}}

\newcommand{\vPhi}{\myvecsym{\Phi}}

\newcommand{\vpsi}{\myvecsym{\psi}}
\newcommand{\vPsi}{\myvecsym{\Psi}}

\newcommand{\vtheta}{\myvecsym{\theta}}

\newcommand{\vSigma}{\myvecsym{\Sigma}}

\newcommand{\va}{\myvec{a}}
\newcommand{\vb}{\myvec{b}}

\newcommand{\ve}{\myvec{e}}
\newcommand{\vf}{\myvec{f}}
\newcommand{\vg}{\myvec{g}}
\newcommand{\vh}{\myvec{h}}

\newcommand{\vk}{\myvec{k}}

\newcommand{\vm}{\myvec{m}}

\newcommand{\vs}{\myvec{s}}

\newcommand{\vu}{\myvec{u}}
\newcommand{\vv}{\myvec{v}}
\newcommand{\vw}{\myvec{w}}

\newcommand{\vx}{\myvec{x}}

\newcommand{\vy}{\myvec{y}}

\newcommand{\vz}{\myvec{z}}

\newcommand{\vA}{\myvec{A}}
\newcommand{\vB}{\myvec{B}}

\newcommand{\vF}{\myvec{F}}

\newcommand{\vH}{\myvec{H}}
\newcommand{\vI}{\myvec{I}}

\newcommand{\vK}{\myvec{K}}

\newcommand{\vP}{\myvec{P}}
\newcommand{\vQ}{\myvec{Q}}
\newcommand{\vR}{\myvec{R}}
\newcommand{\vS}{\myvec{S}}

\newcommand{\vU}{\myvec{U}}
\newcommand{\vV}{\myvec{V}}
\newcommand{\vW}{\myvec{W}}
\newcommand{\vX}{\myvec{X}}

\newcommand{\mymathcal}[1]{\mathcal{#1}}

\newcommand{\calA}{\mymathcal{A}}

\newcommand{\calD}{{\mymathcal{D}}}

\newcommand{\calL}{\mymathcal{L}}
\newcommand{\calM}{\mymathcal{M}}
\newcommand{\calN}{\mymathcal{N}}

\newcommand{\Ga}{\mathrm{Ga}}

\newcommand{\gauss}{\mathcal{N}}

\newcommand{\IG}{\mathrm{IG}}

\newcommand{\MVNIG}{\mathrm{NIG}}

\newcommand{\argmax}{\operatornamewithlimits{argmax}}
\newcommand{\argmin}{\operatornamewithlimits{argmin}}

\newcommand{\half}{\frac{1}{2}}

\newcommand{\defeq}{\triangleq}

\newcommand{\real}{\mathbb{R}}

\newcommand{\trans}{{\mkern-1.5mu\mathsf{T}}}

\newcommand{\ra}{\rightarrow}

\newcommand{\tr}{\mathrm{tr}}

\newcommand{\old}{\mathrm{old}}

\newcommand{\params}{\vtheta}

\newcommand{\nparams}{D}

\newcommand{\relu}{\ensuremath{\mathrm{ReLU}}\xspace}

\newcommand{\expect}[1]{\mathbb{E}\left[{#1}\right]} 
\newcommand{\expectQ}[2]{\mathbb{E}_{{#2}}\left[ {#1} \right]} 

\newcommand{\var}[1]{\mathbb{V}\left[ {#1}\right]}
\newcommand{\varQ}[2]{\mathbb{V}_{{#2}}\left[ {#1}\right]}

\newcommand{\state}{s}

\newcommand{\loss}{\calL}

\newcommand{\data}{\calD}

\newcommand{\NlatentKF}{N_z}
\newcommand{\NobsKF}{N_y}

\newcommand{\hmmhid}{\vz}

\newcommand{\hmmobs}{\vy}

\newcommand{\be}{\begin{equation}}
\newcommand{\ee}{\end{equation}}

\newcommand{\bea}{\begin{eqnarray}}
\newcommand{\eea}{\end{eqnarray}}
\newcommand{\beaa}{\begin{eqnarray*}}
\newcommand{\eeaa}{\end{eqnarray*}}
\newcommand{\ba}{\begin{align*}}
\newcommand{\ea}{\end{align*}}

\newcommand{\figfix}[1]{}

\newcommand{\gitlatex}[2]{} 

\newcommand{\addfig}[4] 
{
    \begin{figure}
    \centering
    \includegraphics[#1]{\figdir/#4}
    \caption{#2}
    \label{fig:#3}
    \end{figure}
}

\newcommand{\addtwofigs}[5]    
{
    \begin{figure}
    \centering
    \begin{subfigure}{0.45\textwidth}
      \centering
      \includegraphics[#1]{\figdir/#4}
      \caption{ }
     \end{subfigure}
~
    \begin{subfigure}{0.45\textwidth}
            \centering
            \includegraphics[#1]{\figdir/#5}
                  \caption{ }
    \end{subfigure}
    \caption{#2}
    \label{fig:#3}
    \end{figure}
}

\newcommand{\addthreefigs}[6]    
{
    \begin{figure}
    \centering
    \begin{subfigure}{0.3\textwidth}
      \centering
      \includegraphics[#1]{\figdir/#4}
      \caption{ }
     \end{subfigure}
~
    \begin{subfigure}{0.3\textwidth}
            \centering
            \includegraphics[#1]{\figdir/#5}
                  \caption{ }
  \end{subfigure}
~
    \begin{subfigure}{0.3\textwidth}
            \centering
            \includegraphics[#1]{\figdir/#6}
                  \caption{ }
    \end{subfigure}
    \caption{#2}
    \label{fig:#3}
    \end{figure}
}

\newcommand{\addfourfigs}[7]    
{
    \begin{figure}
    \centering
    \begin{subfigure}{0.45\textwidth}
      \centering
      \includegraphics[#1]{\figdir/#4}
      \caption{ }
     \end{subfigure}
~
    \begin{subfigure}{0.45\textwidth}
      \centering
      \includegraphics[#1]{\figdir/#5}
      \caption{ }
     \end{subfigure}
\\
    \begin{subfigure}{0.45\textwidth}
      \centering
      \includegraphics[#1]{\figdir/#6}
      \caption{ }
     \end{subfigure}
~
    \begin{subfigure}{0.45\textwidth}
      \centering
      \includegraphics[#1]{\figdir/#7}
      \caption{ }
     \end{subfigure}
    \caption{#2}
    \label{fig:#3}
    \end{figure}
}

\mdfdefinestyle{fearns}{%
    linecolor=red,
    outerlinewidth=50pt,
    roundcorner=20pt,
    innertopmargin=20pt,
    innerbottommargin=20pt,
    innerrightmargin=20pt,
    innerleftmargin=20pt,
    backgroundcolor=yellow!50!white}

\mdfdefinestyle{coptcomment}{%
    linecolor=blue,
    innertopmargin=10pt,
    innerbottommargin=10pt,
    innerrightmargin=10pt,
    innerleftmargin=10pt,
    backgroundcolor=green!25!white}

\mdfdefinestyle{alemi}{%
    linecolor=blue,
    outerlinewidth=50pt,
    roundcorner=20pt,
    innertopmargin=20pt,
    innerbottommargin=20pt,
    innerrightmargin=20pt,
    innerleftmargin=20pt,
    backgroundcolor=blue!30!white}

\newcommand{\action}{a}

\newcommand{\points}[1]{}

\def\vzero{{\bm{0}}}

\def\vmu{{\bm{\mu}}}
\def\vtheta{{\bm{\theta}}}
\def\va{{\bm{a}}}
\def\vb{{\bm{b}}}

\def\ve{{\bm{e}}}
\def\vf{{\bm{f}}}
\def\vg{{\bm{g}}}
\def\vh{{\bm{h}}}

\def\vk{{\bm{k}}}

\def\vm{{\bm{m}}}

\def\vs{{\bm{s}}}

\def\vu{{\bm{u}}}
\def\vv{{\bm{v}}}
\def\vw{{\bm{w}}}
\def\vx{{\bm{x}}}
\def\vy{{\bm{y}}}
\def\vz{{\bm{z}}}

\newcommand{\limtwo}{LiM2}
\newcommand{\updateFreq}{T_u}
\newcommand{\nparamsSub}{d}
\newcommand{\nparamsSmall}{\nparamsSub}
\renewcommand{\nparams}{D}
\newcommand{\nparamsHead}{D_h}
\newcommand{\nparamsBody}{D_b}
\newcommand{\nstates}{N_s}
\newcommand{\ninputs}{N_x}
\newcommand{\nfeatures}{N_z}

\newcommand{\nhidden}{N_h}
\newcommand{\nactions}{N_a}

\newcommand{\npulls}{N_w}
\newcommand{\batchsize}{N_b}
\newcommand{\memory}{M}

\newcommand{\npgd}{N_p}
\newcommand{\nepochs}{N_e}
\newcommand{\fwdcost}{C_f}

\newcommand{\nwarmup}{\tau}
\newcommand{\warmup}{\nwarmup}
\newcommand{\belief}{\vb}
\newcommand{\rewardfn}{f}
\renewcommand{\freward}{\rewardfn}
\renewcommand{\state}{\vs}
\renewcommand{\action}{a}
\newcommand{\reward}{y}

\usepackage[
  style=alphabetic,
  citestyle=alphabetic,
  natbib=true,
  backend=bibtex,
  maxcitenames=3,  mincitenames=1,
  maxbibnames=6,  minbibnames=1,
  firstinits=true,
  backref=true, 
  hyperref=true,
  doi=false,  isbn=false,  url=false,
  arxiv=abs
]{biblatex}

\eat{
\usepackage[
  style=apa,
  citestyle=apa,
  natbib=true,
  backend=biber,
    backref=false,
  hyperref=true,
  doi=false,  isbn=false,  url=false,
  arxiv=abs
]{biblatex}
}

\DefineBibliographyStrings{english}{%
  backrefpage = {page},
  backrefpages = {pages},
}

\usepackage{authblk}

\begin{document}

\eat{
\twocolumn[

\aistatstitle{Efficient Online Bayesian Inference for Neural Bandits}

\aistatsauthor{ Gerardo Duran-Martin \And Aleyna Kara \And  Kevin Murphy }

\aistatsaddress{ Queen Mary University \And Boğaziçi University \And Google Research }
]
}

\title{Efficient Online Bayesian Inference for Neural Bandits}
\author[1]{Gerardo Duran-Martin}
\affil[1]{Queen Mary University, UK}
\author[2]{Aleyna Kara}
\affil[2]{Boğaziçi University, Turkey}
\author[3]{Kevin Murphy}
\affil[3]{Google Research, USA}
\maketitle

\begin{abstract}
In this paper we present a new algorithm for online (sequential) inference
in Bayesian neural networks, and show its suitability for tackling contextual bandit problems. The key idea is to combine the extended Kalman filter (which locally linearizes the likelihood function at each time step) with a (learned or random) low-dimensional affine subspace for the parameters; the use of a subspace  enables us to scale our algorithm to models with $\sim 1M$ parameters. While most other neural bandit methods need to store the entire past dataset in order to  avoid the problem of ``catastrophic forgetting'', our approach uses constant memory. This is possible because we represent uncertainty about all the parameters in the model, not just the final linear layer. We show good results on the ``Deep Bayesian Bandit Showdown'' benchmark, as well as MNIST and a recommender system.
\end{abstract}

\section{Introduction}

Contextual bandit problems
 (see e.g., \citep{Lattimore2019,Slivkins2019})
are a special case of reinforcement learning,
in which the state (context) at each time step is chosen independently,
rather than being dependent on the past history of states and actions.
Despite this limitation,
contextual bandits
are widely used in real-world applications,
such as
recommender systems \citep{Li10linucb,Guo2020bandits},
advertising \citep{McMahan13,Du2021kdd},
healthcare \citep{Greenewald2017,Aziz2021},
etc.
The goal is to maximize the sequence of rewards $\reward_t$
obtained by picking actions $\action_t$ in response
to each input context or state $\state_t$.
To do this, the decision making agent
must learn a reward model $\expect{\reward_t|\state_t,\action_t,\vtheta} = \freward(\state_t,\action_t;\vtheta)$,
where $\vtheta$ are the unknown model parameters.
Unlike supervised learning, the agent does not get to see
the ``correct'' output, but instead only
gets feedback on whether the choice it made was good or bad
(in the form of the reward signal).
If the agent knew $\vtheta$, it could pick the optimal
action using
$\action_t^* = \argmax_{\action \in \calA} \freward(\state_t,\action;\vtheta)$.
However, since $\vtheta$ is unknown, the agent must ``explore'',
so it can gather information about the reward function,
before it can ``exploit'' its model.

In the bandit literature,
the two most common solutions to solving the explore-exploit dilemma
are based on the upper confidence bound (UCB) method
(see e.g., \citep{Li10linucb,Kaufmann2012})
and the Thompson Sampling (TS) method (see e.g., \citep{Agrawal2013icml,Russo2018}).
The key bottleneck in both UBC and TS is efficiently computing the posterior
$p(\vtheta|\data_{1:t})$ in an online fashion,
where $\data_{1:t}=\{(\state_i,\action_i,\reward_i): i=1:t\}$
is all the data seen so far.
This can be done in closed form for linear-Gaussian models,
but for nonlinear models, such as deep neural networks (DNNs), it is computationally infeasible.

In this paper, we propose to use
a version of the extended Kalman filter
to recursively approximate the parameter posterior 
$p(\vtheta|\data_{1:t})$ using constant time and memory
(i.e., independent of $T$).
The main novelty of our approach is that
we show how to scale the EKF to large neural networks
by leveraging recent results that show that deep neural networks often
have very few ``degrees of freedom''
(see e.g.,
\citep{Li2018Intrinsic,Izmailov2019,Larsen2021degrees}).
Thus we can  compute  a low-dimensional
subspace
and perform Bayesian filtering in the subspace
rather than the original parameter space.
We therefore call our method ``Bayesian subspace bandits''.

Although Bayesian inference in DNN subspaces has previously
been explored (see related work in \cref{sec:related}),
it has not been done in an online or bandit setting, as far as we know.
Since we are using approximate inference,
we lose the well-known optimality
of Thompson sampling \citep{Phan2019};
we leave proving regret bounds for our method to future work.
In this paper, we restrict attention to an empirical comparison.
We show that our method works well in practice
on various datasets,
including the ``Deep Bayesian Bandits Showdown'' benchmark
\citep{Riquelme2018},
the MNIST dataset, and a recommender system dataset.
In addition, our method uses much less memory and time
than most other methods.

Our algorithm is not specific to bandits,
and can be  applied to any situation that requires efficient online computation
of the posterior. This includes tasks such as life long learning,
Bayesian optimization,
active learning, reinforcement learning, etc.\footnote{
These problems are all very closely related.
For example, BayesOpt is a kind of (non-contextual) bandit problem
with an infinite number of arms;
the goal is to identify the action (input to the reward
function $f: \real^D \ra \real$)
that maximizes the output.
Active learning is closely related to BayesOpt, but now the actions correspond to
choosing data points $\vx \in \real^n$ that we want to label, and our
objective is to minimize uncertainty about the underlying function $f$,
rather than find the location of its maximum.
}.
However, we leave such extensions to future work.


\section{Related work}
\label{sec:related}

In this section, we briefly review related work.
We divide the prior work into several groups:
Bayesian neural networks,
neural net subspaces,
and neural contextual bandits.

Most work on Bayesian inference for neural networks
has focused on the offline (batch) setting.
Common approaches include
the Laplace approximation
\citep{MacKay1992, MacKay95,Daxberger2021laplace};
Hamiltonian MCMC
\citep{neal1995bayesian,Izmailov2021icml};
variational inference,
such as the ``Bayes by backprop'' method of
\citep{Blundell2015},
and the
``variational online Gauss-Newton''
method of \citep{Osawa2019nips};
expectation propagation,
such as the
``probabilistic backpropagation'' method of
\citep{PBP};
and many others.
(For more details and references,
see e.g., \citep{Polson2017,Wilson2020BDL,Wilson2020prob,Khan2020tutorial}.)

There are several techniques for online or sequential Bayesian inference
for  neural networks.
\citep{Ritter2018online} propose an online version of the Laplace
approximation, \citep{Nguyen2018vcl}
propose an online version of variational inference,
and \citep{Ghosh2016} propose
to use assumed density filtering (an online version of expectation
propagation).
However, in \citep{Riquelme2018}, they showed that these methods
do not work very well for bandit problems.
In this paper, we build on older work, specifically
\citep{Singhal1988,deFreitas00ekf},
which used the extended Kalman filter (EKF)
to perform approximate online inference for DNNs.
We combine this with subspace methods to scale to high dimensions,
as we discuss below.

There are several techniques for scaling Bayesian inference to neural
networks with many parameters.
A simple approach is to use variational inference with a diagonal
Gaussian posterior, but this ignores important correlations between
the weights.
It is also possible to use low-rank factorizations of the posterior
covariance matrix.
In \citep{Daxberger2021subnetwork},
they propose to use a MAP estimate for some parameters
and a Laplace approximation for others.
However, their computation of the MAP estimate relies on standard offline SGD (stochastic gradient descent),
whereas we perform online Bayesian inference without using SGD.
\eat{
they propose to partition the weights $\vtheta \in \real^D$
into a high dimensional set, $\vtheta_1 \in \real^{D-d}$,
and a low dimensional set, $\vtheta_2 \in \real^{d}$.
To find this partition, they compute a diagonal Laplace approximation
to $p(\vtheta|\data)$, and then select the $d$ elements of $\vtheta$
with highest posterior marginal variance.
They then compute a full covariance Laplace approximation
over $\vtheta_2$, treating $\vtheta_1$ as point estimate.
This gives the following posterior approximation:
$p(\vtheta|\data) \approx \delta(\vtheta_1 - \hat{\vtheta}_1)
\gauss(\vtheta_2 | \hat{\vtheta}_2, \vSigma_2)$,
where $\hat{\vtheta}=(\hat{\vtheta}_1,\hat{\vtheta}_2) \in \real^D$ is the usual MAP estimate,
and $\vSigma_2$ is the inverse of the Hessian for $\vtheta_2$ at the posterior mode.
}
In \citep{Izmailov2019}, they compute
a linear subspace of dimension $\nparamsSub$
by applying PCA to the last $L$ iterates
of stochastic weight averaging
\citep{Izmailov2018};
they then perform slice sampling in this low-dimensional subspace.
In this paper, we also leverage subspace inference,
but we do so in the online setting, which is
necessary when  solving bandit problems.

The literature on contextual bandits is vast
(see e.g., \citep{Lattimore2019,Slivkins2019}).
Here we just discuss  recent work which utilizes DNNs
to model the reward function,
combined with Thompson sampling as the policy for choosing the action.
In \citep{Riquelme2018}, they evaluated
many different approximate inference methods for Bayesian neural networks
on a set of benchmark contextual bandit problems;
they called this the ``Deep Bayesian Bandits Showdown''.
The best performing method
in their showdown 
is what they call the ``neural linear'' method,
which we discuss in \cref{sec:neuralLinear}.

Unfortunately the neural linear method is not a fully online algorithm,
since it needs to keep all the past data 
to avoid the problem of  ``catastrophic forgetting''
\citep{Robins1995,French99,Kirkpatrick2017}.
This means that the memory complexity is $O(T)$,
and the computational complexity can be as large as $O(T^2)$.
This makes the method impractical for applications
where the data is high dimensional,
and/or the agent is running for a long time.
In \citep{Nabati2021}, they make an online version of the neural linear
method which they  call "Lim2", which stands for
``Limited Memory Neural-Linear with Likelihood Matching''.
We discuss this in more detail in \cref{sec:LIM}.

More recently, several methods based on neural tangent kernels (NTK)
have been developed \citep{NTK},
including
neural Thompson sampling \citep{neuralTS}
and neural UCB  \citep{neuralUCB}.
We discuss these methods in more detail in \cref{sec:neuralTS}.
Although Neural-TS and Neural-UCB in principle
achieve a regret of $O(\sqrt{T})$, in practice there are some
disadvantages.
First, these algorithms
perform multiple gradient steps, based on all the past data,
at each step of the algorithm.
Thus these are full memory algorithms that take $O(T)$ space
and $O(T^2)$ time.
Second,
it can be shown \citep{Allen-Zhu2019,Ghorbani2020}
that NTKs are less data efficient learners than (finite width)
hierarchical DNNs, both in theory and in practice.
Indeed we will show that our approach, that uses
constant memory and  finite width DNNs,
does significantly better in practice.

\eat{
\begin{itemize}

\item \citep{Daxberger2021subnetwork}
  ``Bayesian Deep Learning via Subnetwork Inference''.
  
\item \citep{Izmailov2019}
  ``Subspace Inference for Bayesian Deep Learning''. PCA on iterates
  after SGD converges.
  
\item \citep{Nguyen2018vcl}
``Variational Continual Learning''.

\item \citep{Ritter2018online}
  ``Online Structured Laplace Approximations for Overcoming
  Catastrophic Forgetting''.

\item \citep{Daxberger2021laplace}.
  ``Laplace Redux--Effortless Bayesian Deep Learning''.

\item  \citep{Ghosh2016} ADF for DNNs.
  
\item \citep{Urteaga2017}, they also use mixture models 
to model the reward function for Thompson sampling,
but they do not use neural networks (so they cannot handle
high dimensional input contexts),
and they use variational inference rather than EKF.

\item \citep{Sezener2020}.
"Online learning in contextual bandits using Gated Linear
               Networks".

\end{itemize}
}

\section{Methods}
\label{sec:methods}

In this section, we discuss various methods for tackling
bandit problems, including our proposed new method.

\eat{
  https://github.com/ZeroWeight/NeuralTS
  
https://github.com/SMPyBandits/SMPyBandits/blob/master/NonStationaryBandits.md

https://github.com/david-cortes/contextualbandits

https://github.com/st-tech/zr-obp
}

\subsection{Algorithmic framework}

\begin{algorithm}
\caption{Online-Eval(Agent, Env, $T$, $\nwarmup$)}
\label{algo:generic}
\footnotesize
\dontprintsemicolon
$\data_{\nwarmup} = \text{Environment.Warmup}(\nwarmup)$\;
$\belief_{\nwarmup} = \text{Agent.InitBelief}(\data_{\nwarmup})$ \;
$R = 0$ // cumulative reward \;
\For{$t=(\nwarmup+1):T$}{
$\state_t = \text{Environment.GetState}(t)$ \;
$\action_{t} = \text{Agent.ChooseAction}(\belief_{t-1}, \state_t)$ \;
  $\reward_t = \text{Environment.GetReward}(\state_t,\action_t)$ \;
  $R += \reward_t$ \;
 $\data_t = (\state_t, \action_t, \reward_t)$ \;
  $\belief_{t} = \text{Agent.UpdateBelief}(\belief_{t-1}, \data_t)$\;
}
Return $R$
\end{algorithm}

In \cref{algo:generic},
we give the pseudocode for a way to estimate the expected
reward for a bandit policy (agent),
given access to an environment or simulator.
In the case of a Thompson sampling agent,
the action selection  is usually implemented
by first sampling a parameter vector from the posterior (belief state),
  $\tilde{\params}_{t} \sim p(\params_{t}|\data_{1:t-1})$,
and then predicting the reward for each action and greedily picking the best,
  $\action_t = \argmax_{\action \in \calA}
  \expect{\reward| \state_t,\action_t, \tilde{\params}_t}$.
In the case of a UCB agent, the action is chosen by first computing
the posterior predicted mean and variance,  and then picking the
action with the highest optimistic estimate of reward:
\begin{align}
  p_{t|t-1}(\reward|\state, \action) &\defeq
  \int p(\reward|\state,\action, \params)  p(\params|\data_{1:t-1}) d\params \\
  \mu_{a} &=   \expectQ{\reward|\state_t,\action}{p_{t|t-1}} \\
  \sigma_a &= \sqrt{\varQ{\reward|\state_t,\action}{p_{t|t-1}}} \\
  \action_t &= \argmax_{\action \in \calA} \mu_a + \alpha \sigma_a
\end{align}
where $\alpha>0$ is a tuning parameter that controls the degree of exploration.
In this paper, we focus on Thompson sampling,
but our methods can be extended to UCB in a straightforward way.

Since the prior on the parameters is usually uninformative,
the initial actions are effectively random.
Consequently we let the agent have a ``warmup period'',
in which we systematically try each action $\npulls$ times, in a round robin fashion,
for a total of $\nwarmup=\nactions \times \npulls$ steps.
We then use this warmup data
 to initialize the belief state to get an informative prior.
If we have a long warmup period,
then we will have a better initial estimate,
but we may incur high regret during this period,
since we are choosing actions ``blindly''.
Thus we can view $\nwarmup$ as a hyperparameter of the algorithm.
The optimal value
 will depend on the expected lifetime $T$ of the agent
(if $T$ is large, we can more easily amortize the cost of a long warmup period).

\subsection{Modeling assumptions}

\begin{figure*}
  \centering
  \footnotesize
\begin{subfigure}[b]{0.3\textwidth}
\centering
\includegraphics[height=1.2in]{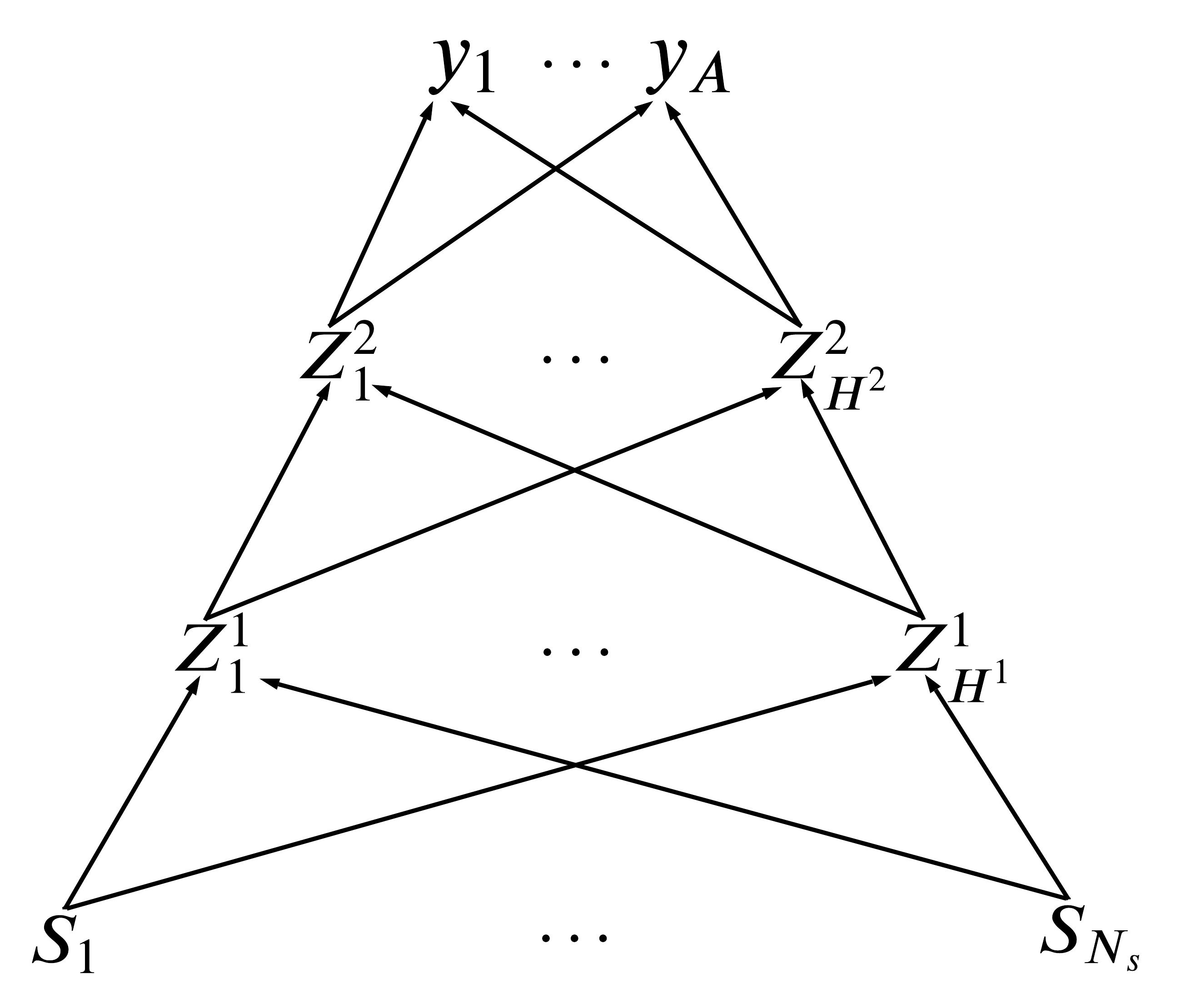}
\caption{ }
\label{fig:mlp-multi-head}
\end{subfigure}
~
\begin{subfigure}[b]{0.3\textwidth}
  \centering
  \includegraphics[height=1.2in]{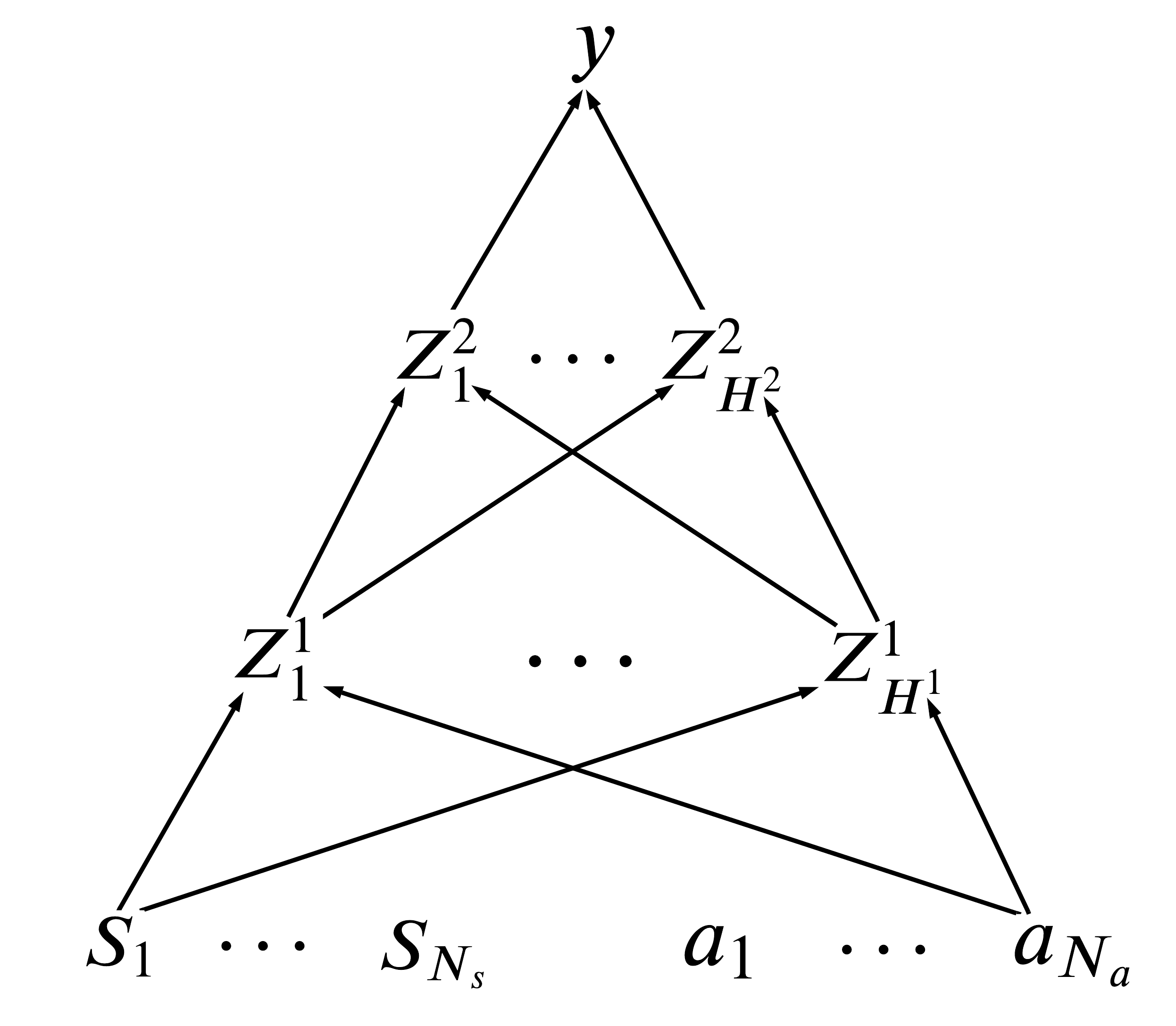}
  \caption{ }
  \label{fig:mlp-concat}
\end{subfigure}
~
\begin{subfigure}[b]{0.3\textwidth}
  \centering
  \includegraphics[height=1.2in]{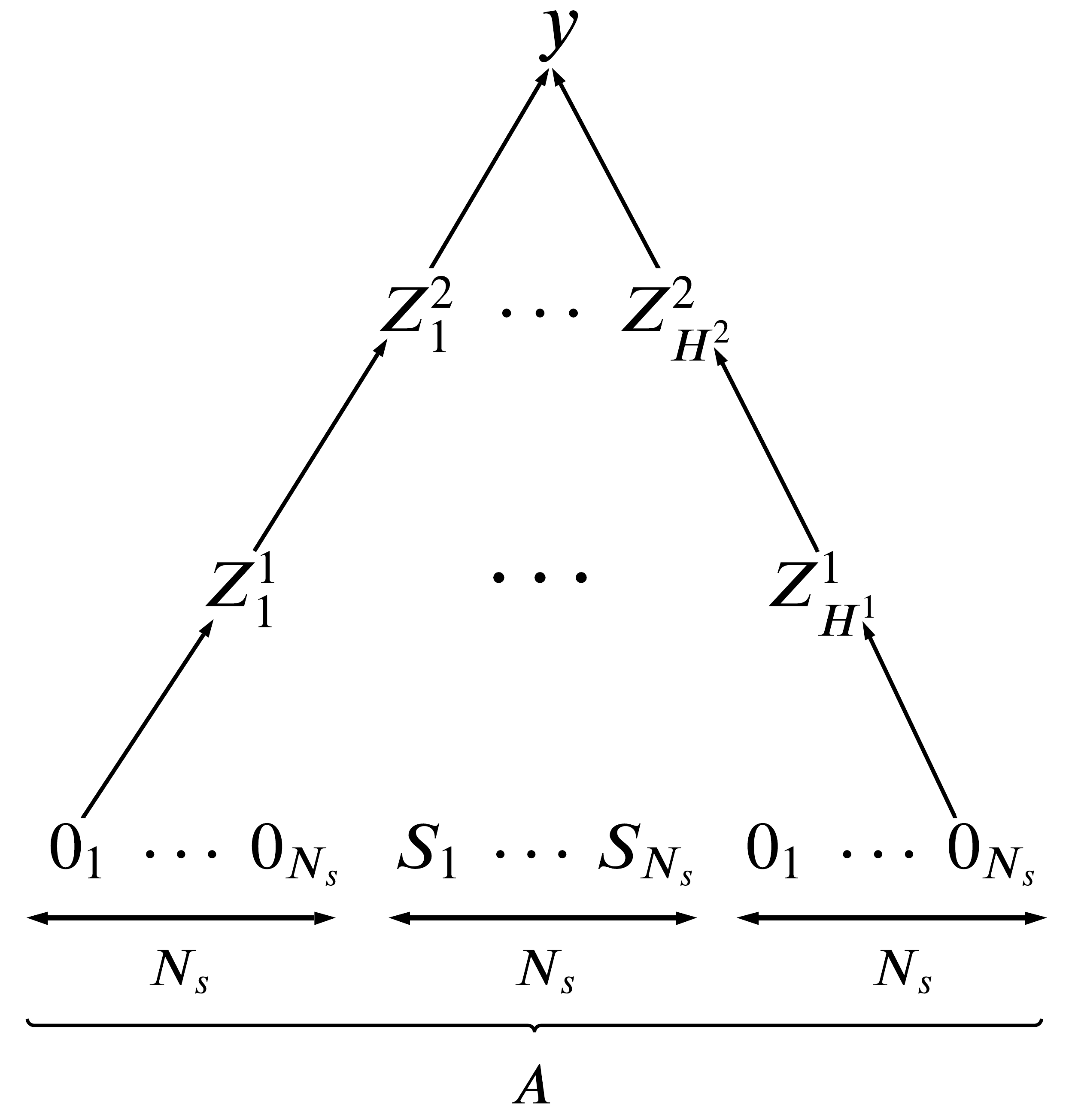}
  \caption{ }
  \label{fig:mlp-repeat}
\end{subfigure}
\caption{
  Illustration of some common MLP architectures
  used in bandit problems.
  $\vs$ represents the state (context) vector,
  $\va$ represents the action vector,
  $\vy$ represents the reward vector (for each possible action),
  and $z_i^l$ is the $i$'th hidden node in layer $l$.
  (a) The input is $\vs$,
 and there are $A$ output ``heads'', $y_1,\ldots,y_A$, one per action.
 (b) The input is a concatentation of $\vs$ and $\va$;
the output is the predicted reward for this $(\vs,\va)$ combination.
(c) The input is a block structured vector,
where we insert $\vs$ into the $a$'th block (when evaluating action $a$),
and the remaining input blocks are zero.
}
\label{fig:MLP}
\end{figure*}

We will assume a Gaussian bandit setting,
in which the observation model for the reward
is a Gaussian
with a fixed or inferred observation variance:
$p(\reward_t|\state_t,\action_t) = \gauss(\reward_t|\rewardfn(\state_t,\action_t;\params_t),\sigma^2)$.
(We discuss extensions to the Bernoulli bandit case in \cref{sec:discuss}.)

Many current bandit algorithms  assume the reward function
is a linear model applied to a set of learned features.
That is, it 
has the form
$f(\state,\action;\params) = \vw_{\action}^\trans \vphi(\state;\vV)$,
where $\vphi(\state;\vV) \in \real^{\nfeatures}$ is the hidden state computed by a feature
extractor, $\vV \in \real^{\nparamsBody}$
are the parameters of this feature extractor ``body'',
and $\vW \in \real^{\nfeatures \times \nactions}$ is the final linear layer,
with one output ``head'' per action.
For example, in \cref{fig:mlp-multi-head},
we show a 2 layer model 
where $\vphi(\state;\vV) = \relu(\vV_2 \; \relu(\vV_1 \state))$
is the feature vector,
and $\vV_1 \in \real^{\nhidden^{(1)} \times \nstates}$
and $\vV_2 \in \real^{\nhidden^{(2)} \times \nhidden^{(1)}}$
are the first and second layer weights.
(We ignore the bias terms for simplicity.)
Thus $\nfeatures=\nhidden^{(2)}$ is the size of the feature vector
that is passed to the final linear layer.
If the feature vector is fixed (i.e., is not learned),
so $\vphi(\vs)=\vs$,
we get a linear model
of the form
$f(\state,\action;\vw) = \vw_{\action}^\trans \state$.

An alternative model structure is to
concatentate the state vector, $\vphi(\state_t)$,
with the action vetcor, $\vphi(\action_t)$ 
to get an input of the form
$\vx_t=(\vphi(\state_t),\vphi(\action_t))$.
This is shown in \cref{fig:mlp-concat}.
This can be useful if we have many possible actions;
in this case, we can represent arms in terms of their features instead
of their indices, just as we represent states in terms of their features.
In this formulation,  the linear output layer returns
the predicted reward for the specified $(\vs,\va)$ input combination,
and we require $\nactions$ forwards passes to evaluate the reward vector
for each possible action.

Instead of concatenating the state and action vectors,
we can compute their outer product and then flatten the result,
to get $\vx_t = \text{flatten}(\vphi(\state_t) \vphi(\action_t)^\trans)$.
This can model interaction effects,
an proposed in
\citep{Li10linucb}.
If $\vphi(\action_t)$ is a one-hot encoding,
we get the block-structured input
$\vx_t = (\vzero, \cdots, \vzero, \vphi(\state_t), \vzero, \cdots, \vzero)$,
where we insert the state feature vector into the
block corresponding to the chosen action
(see \cref{fig:mlp-repeat}).
This approach is used by recent NTK methods.
If we  assume $\vphi(\vs)=\vs$, so the state features are fixed,
and we assume that the MLP has no hidden layers,
then this model becomes equivalent to the linear model,
since $\vw^\trans \vx_t = \vw_a^\trans \state_t$.

\subsection{Existing methods}

In this section, we briefly describe existing inference methods that we will compare to.
More details on all methods can be found in the Supplementary Information.
These methods differ in  the kind of belief state
they use to represent uncertainty about the model parameters,
and in their mechanism for updating this belief state.
See \cref{tab:methods} for a summary.

\begin{table*}
  \centering
 \footnotesize
  \begin{tabular}{llll }
    Method & Belief state & Memory & Time \\ \hline
    Linear & $(\vmu_{t,a},\vSigma_{t,a})$
    & $O(\nactions \nfeatures^2)$
    & $O(T (\fwdcost + \nactions \ninputs^3))$
    \\
    Neural-Greedy & $\vtheta_t=(\vV_t,\vW_t)$
    & $O(\nparamsBody + \nactions \nfeatures + T \ninputs)$
    &$O(T' T \nepochs \fwdcost)$
    \\
    Neural-Linear & $(\vV_t, \vmu_{t,a},\vSigma_{t,a}, \data_{1:t})$
    & $O(\nparamsBody + \nactions \nfeatures^2 + T  \ninputs)$
    & $O(T' T \nepochs \fwdcost + T \nactions \nfeatures^3)$
    \\
        LiM2 & $(\vV_t, \vmu_{t,a}, \vSigma_{t,a}, \data_{t-\memory:t})$
    & $O(\nparamsBody + \nactions \nfeatures^2  + \memory  \ninputs)$
    & $O(T'  \memory \nepochs \npgd(\fwdcost + \nfeatures^3)     + T \nactions \nfeatures^3)$
    \\
    Neural-Thompson & $(\vtheta_t, \vSigma_{t}, \data_{1:t})$
    & $O(\nparams + \nparams^2  + T  \ninputs)$
    & $O(T (T \nepochs \fwdcost + \nparams^3))$
    \\
    EKF & $(\vmu_t,\vSigma_t)$
    & $O(\nparams^2)$
    & $O(T (\fwdcost + \nparams^3))$
    \\
    EKF-Subspace & $(\vmu_t,\vSigma_t,\vtheta_*,\vA)$
    & $O(\nparamsSmall^2 + \nparams \nparamsSmall)$
    & $O(T (\fwdcost +  \nparamsSmall^3 +  \nparams \nparamsSmall))$ 
  \end{tabular}
  \caption{Summary of the methods for Bayesian inference
    considered in this paper.
    Notation:
    $T$: num steps taken by the agent in the environment;
    $\updateFreq$: update frequency for SGD;
    $T' =T / \updateFreq$: total num.  times     that we invoke SGD;
    $\nepochs$: num. epochs over the training data for each run of SGD;
    $\fwdcost$: cost of to evaluate gradient of the network on one example,
    $\nactions$: num.  actions;
    $\ninputs$: size of input feature vector for state and action.
    $\nfeatures$: num.  features in  penultimate (feature) layer;
    $\nparamsBody$: num.  parameters in the body (feature extractor);
    $\nparamsHead = \nactions \nfeatures$: num.  parameters in final layer linear;
    $\nparams=\nparamsBody + \nparamsHead$: total num.  parameters;
    $\nparamsSmall$: size of subspace;
    $\memory$: size of memory buffer;
    }
  \label{tab:methods}
  \end{table*}

\paragraph{Linear method}
\label{sec:linear}

The most common approach to bandit problems is to assume a linear model for
the expected reward,
$f(\state,\action;\params) = \vw_{\action}^\trans \state$.
If we use a Gaussian prior, and assume a Gaussian likelihood,
then we can
represent the belief state  as a Gaussian,
$\belief_t = \{(\vmu_{t,a}, \vSigma_{t,a}): a=1:\nactions\}$.
This can be efficiently
updated online using the recursive least squares algorithm,
which is a special case of the Kalman filter
(see \cref{app:linearBandit} for details).

\paragraph{Neural linear method}
\label{sec:neuralLinear}

In  \citep{Riquelme2018},
they proposed a method called
``neural linear'', which they showed outperformed many other more sophisticated approaches,
such as variational inference,
on their bandit showdown benchmark.
It assumes that the reward model has the form
$f(\state,\action;\params) = \vw_{\action}^\trans \vphi(\state;\vV)$,
where $\vphi(\state;\vV) \in \real^{\nfeatures}$ is the hidden state computed by a feature
extractor 
(see \cref{fig:mlp-multi-head} for an illustration).
The neural linear method computes
a point estimate of $\vV$ by using SGD,
and uses Bayesian linear regression to update the posterior
over each $\vw_a$, and optionally $\sigma^2$.

If we just update $\vV$ at each step using $\data_t$,
we run the risk of
``catastrophic forgetting'' (see \cref{sec:related}).
The standard solution to this
is to store all the past data,
and to re-run (minibatch) SGD on all the data at each step.
Thus the belief state is represented as
$\belief_t = (\vtheta_t, \data_{1:t})$.
See \cref{app:neuralLinear} for details.

The time cost is $O(T^2 \nepochs \fwdcost)$,
where  $\nepochs$ is the number of epochs (passes over the data) at each step,
and  $\fwdcost$ is the cost of a single forwards-backwards pass
through the network (needed to compute the objective and its gradient).\footnote{
The reason for the quadratic cost is that each epoch
passes over $O(T)$ examples, even if we use minibatching.
} 
Since it is typically  too expensive to run SGD on each step,
we can just perform updating every $\updateFreq$ steps.
The total time then becomes
$O(T' T  \nepochs \fwdcost)$,
where $T'=T/\updateFreq$ is the total number of times we invoke SGD.

The memory cost is
$O(\nparams + T \ninputs)$,
where $\ninputs$ is the size of
each input example, $\vx_t=(\state_t, \action_t)$.
If we limit the memory to the last $\memory$ observations
(also called a ``replay buffer''),
the memory reduces to $O(\nparams + \memory \ninputs)$,
and the time reduces to
$O(T' \memory \nepochs  \fwdcost)$.
However, naively limiting the memory in this way can hurt (statistical) performance,
as we will see.

\paragraph{\limtwo}
\label{sec:LIM}

In \citep{Nabati2021},
they propose a method called ``LiM2'', which stands for
``Limited Memory Neural-Linear with Likelihood Matching''.
This is an extension of the neural
linear method 
designed to solve the ``catastrophic forgetting''
that occurs when using a fixed memory buffer.
The basic idea is to approximate the covariance
of the old features in the memory buffer before
replacing them with the new features,
computed after updating the network parameters.
This old covariance can be used as  a prior
during the Bayesian linear regression step.

Computing the updated prior covariance requires
solving a semi-definite program (SDP) after each SGD step.
In practice, the SDP can be solved using
an inner loop of projected gradient descent (PGD),
which involves solving an eigendecomposition at each step.
This takes $O(T'  \memory \nepochs \npgd(\fwdcost + \nfeatures^3))$ time,
where $\npgd$ is the number of PGD steps per SGD step.
See \cref{app:LIM2} for details.

\paragraph{NTK methods}
\label{sec:neuralTS}

In \citep{neuralTS}, they propose a method called
``Neural Thompson Sampling'',
and in \citep{neuralUCB},
the propose a related method called
``neural UCB''. Both methods are based on approximating
the MLP with a neural tangent kernel or NTK \citep{NTK}.
Specifically, the feature vector at time $t$ is defined
to be
$\vphi_t(\state,\action) = (1/\sqrt{\nhidden}) \nabla_{\vtheta} \freward(\state,\action)|_{\vtheta_{t-1}}$,
where $\nhidden$ is the width of each hidden layer,
and the gradient is evaluated at the most recent parameter estimate.
They use a linear Gaussian model on top of these features.
The network parameters are re-estimated at each step based on all the past data,
and then the method
effectively performs Bayesian linear regression on the output layer
(see \cref{app:neuralTS} for details).

\subsection{Our method: Subspace EKF}
\label{sec:subspace}
\label{sec:EKF}

A natural alternative to just modeling uncertainty
in the final layer weights is to ``be Bayesian'' about {\em all} the network parameters.
Since our model is nonlinear,
we must use approximate Bayesian inference.
In this paper we choose to use
the Extended Kalman Filter (EKF),
which is a popular
deterministic inference scheme for nonlinear state-space models
based on linearizing the model
(see \cref{app:EKF} for details).
It was first applied to inferring the parameters of an MLP in
\citep{Singhal1988},
although it has not been applied to bandit problems, as far as we know.
In more detail, we  define the latent variable
to be the unknown parameters $\vtheta_t$.
The (non-stationary) observation model is given by
$p_t(\reward_t|\vtheta_t) =
\gauss(\reward_t|\rewardfn(\state_t,\action_t;\vtheta_t,\sigma^2)$,
where $\state_t$ and $\action_t$ are inputs to the model,
and the dynamics model for the parameters is given by
$p(\vtheta_t|\vtheta_{t-1}) = \gauss(\vtheta_t|\vtheta_{t-1},\tau^2 \vI)$.
We can set $\tau^2 = 0$ to encode the assumption that the
parameters of the reward function are constant over time.
However in practice we use a small non-zero value for $\tau$, for numerical stability.

The belief state of an EKF has the form
$\belief_t = (\vmu_t, \vSigma_t)$.
This takes $O(\nparams^2)$  space
and $O(T \nparams^3)$  time to compute.
Modern neural networks often have millions of parameters,
which makes direct application of the EKF intractable.
We can reduce the memory from $O(\nparams^2)$ to $O(\nparams)$ and the time
from $O(T \nparams^3)$ 
to $O(T \nparams^2)$ by using a diagonal approximation to $\vSigma_t$.
However, this ignores  correlations between the parameters,
which is important for good performance (as we show empirically in \cref{sec:results}).
We can improve the approximation by using a block structured approximation,
with one block per layer of the MLP, but this still ignores correlations
between layers.

In this paper, we explore a different approach to scaling the EKF
to large neural networks.
Our key insight is to exploit the fact that
the DNN  pararameters
are not independent ``degrees of freedom''.
Indeed,
\citep{Li2018Intrinsic}
showed empirically that
we can replace the original neural network weights
$\vtheta \in \real^{\nparams}$ with
a lower dimensional version, $\vz \in \real^{\nparamsSmall}$,
by defining the affine mapping
$\vtheta(\vz) = \vA \vz + \vtheta_*$,
and then optimizing the low-dimensional
parameters $\vz$.
Here $\vA \in \real^{\nparams \times \nparamsSmall}$ is a fixed but random Gaussian matrix
with columns normalized to 1,
and $\vtheta_* \in \real^{\nparams}$ is a random initial guess of the parameters
(which we call an ``offset'').
In \citep{Li2018Intrinsic},
they show that optimizing in the $\vz$ subspace  gives good
results on standard classification and RL benchmarks,
even when $\nparamsSmall \ll \nparams$,
provided that $\nparamsSmall >  \nparamsSmall_{\min}$,
where $\nparamsSmall_{\min}$ is a critical threshold.
In \citep{Larsen2021degrees}, they provide a theoretical
explanation for why such a threshold exists,
based on geometric properties of the high dimensional loss landscape.

Instead of using a random offset $\vtheta_*$, 
we can optimize it by performing  SGD in the original
$\vtheta$ space during a warmup period.
Similarly, instead of using a random basis matrix $\vA$,
we can optimize it
by applying SVD 
to the iterates of SGD during the warmup period,
as proposed in \citep{Izmailov2019,Larsen2021degrees}.
(If we wish, we can   just  keep a subset of the iterates, since consecutive
samples are correlated.)
These two changes reduce the dimensionality of the subspace $\nparamsSmall$
that
we need to use in order to get good performance.
(We can use cross-validation on the data from the warmup phase
to find a good value for $\nparamsSmall$.)

\begin{figure}
\centering
\includegraphics[height=2in]{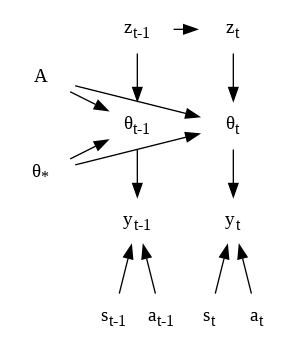}
\caption{
  Graphical model for the subspace bandit.
}
\label{fig:SSM}
\end{figure}

Once we have computed the subspace, we can perform
Bayesian inference for the embedded parameters $\vz \in \real^d$
instead of the original parameters $\vtheta \in \real^D$.
We do this by applying the EKF to the
a state-space model with a (non-stationary) observation model
of the form
$p_t(\reward_t|\vz_t) =
\gauss(\reward_t|\rewardfn(\state_t,\action_t;\vA \vz_t + \vtheta_*),\sigma^2)$,
and a deterministic transition model of the form
$p(\vz_t|\vz_{t-1}) = \gauss(\vz_t|\vz_{t-1}, \tau^2 \vI)$.
This is illustrated as a graphical model in 
\cref{fig:SSM}.

\eat{
Our overall policy is a two-stage method,
in which we first choose actions uniformly for a few steps (as is standard),
and then we switch to Thompson sampling,
combined with our subspace EKF method.
}

The overall algorithm is summarized in \cref{algo:subspaceEKF}.
(If we use a random subspace, we can skip the warmup phase,
but results are worse, as we show in \cref{sec:results}.)
The algorithm takes $O(\nparamsSmall^3)$ time per step.
Empirically we find that we can reduce
models with $\nparams \sim 10^6$ down to
$\nparamsSmall \sim 10^2$ while getting the same (or sometimes better)
performance, as we show in \cref{sec:results}.
We can further reduce the time to $O(\nparamsSmall)$
by using a diagonal covariance matrix, with little change to the performance,
as we shown in \cref{sec:results}.
The time cost of the warmup phase is dominated by SVD.
If we have $\warmup$ samples,
the time complexity  for exact SVD is $O(\min(\warmup^2 \nparams, \nparams^2 \warmup))$.
However, if
we use randomized SVD \citep{Halko2011}
this  reduces the time  to
$O(\warmup \nparams \log \nparamsSmall + (\warmup+\nparams) \nparamsSmall^2)$.

The memory cost is
$O(\nparamsSmall^2 + \nparams \nparamsSmall)$,
since we need to store the belief state,
$\belief_t = (\vmu_t, \vSigma_t)$,
as well as the offset $\vtheta_*$
and the $D \times d$ basis matrix $\vA$.
We have succesfully scaled this to models with $\sim 1M$ parameters,
but going beyond this may require the use of a sparse random
orthogonal matrix to represent $\vA$ \citep{Choromanski2017}.
We leave this to future work.

Note that our method can be applied to any kind of DNN,
not just MLPs. The low dimensional vector $\vz$
depends on all of the parameters in the model.
By contrast, the neural linear and Lim2
methods assume that the model has a linear final layer,
and they only capture parameter uncertainty in this final layer.
Thus these methods cannot be combined with the subspace trick.

\eat{
We find that we can capture 99\% of the variance 
using $\nparamsSmall \sim 5$ dimensions, but we get better end-to-end performance
using a value of $\nparamsSmall \sim 200$.
(The estimate of the dimensionality
based on the warmup data is so low
because the initial iterates only explore a small portion
of parameter space, near to the random initialization.)
}

\begin{algorithm}
\caption{Neural Subspace Bandits}
\label{algo:subspaceEKF}
\dontprintsemicolon
\footnotesize
$\data_{\nwarmup} = \text{Environment.Warmup}(\nwarmup)$\;
$\vtheta_{1:\nwarmup} = \text{SGD}(\data_{\warmup})$ \;
$\vtheta_* = \vtheta_{\nwarmup}$ \;
$\vA = \text{SVD}(\vtheta_{1:\nwarmup})$\;
  $(\vmu_{\nwarmup}, \vSigma_{\nwarmup}) = \text{EKF}(\vmu_{0},\vSigma_{0}, \data_{1:\nwarmup})$\;
\For{$t=(\nwarmup+1):T$}{
  $\state_t = \text{Environment.GetState}(t)$ \;
  $\tilde{\vz}_t \sim \gauss(\vmu_t, \vSigma_t)$ \\
  $\action_t = \argmax_a \rewardfn(\state_t,\action_t;\vA \tilde{\vz}_t + \vtheta_*)$ \;
 $\reward_t = \text{Environment.GetReward}(\state_t,\action_t)$ \;
 $\data_t = (\state_t, \action_t, \reward_t)$ \;
  $(\vmu_{t}, \vSigma_{t}) = \text{EKF}(\vmu_{t-1},\vSigma_{t-1}, \data_t)$\;
}
\end{algorithm}

\section{Results}
\label{sec:results}

\eat{
  https://github.com/sauxpa/neural_exploration

  https://github.com/fidelity/mabwiser
  https://raw.githubusercontent.com/fidelity/mabwiser/master/examples/lints_reproducibility/movielens_responses.csv
  
  https://www.kaggle.com/prajitdatta/movielens-100k-dataset
  100,000 ratings (1-5) from 943 users on 1682 movies.
* Each user has rated at least 20 movies.
* Simple demographic info for the users (age, gender, occupation, zip)

}

In this section, we present empirical results
in which we evaluate the performance (reward) and speed (time)
of our method compared to other methods on various bandit problems.
We also study the effects of various hyper-parameters of our algorithm,
such as how we choose the subspace.

\subsection{Tabular datasets}
\label{sec:tabular}

To compare ourselves to prior works, we consider a subset
of the datasets used in the ``Deep Bayesian Bandits Showdown''
\citep{Riquelme2018}.
These are small tabular datasets, where the goal
is to predict the class label given the features.\footnote{
The datasets are from the UCI ML repository
\protect\url{https://archive.ics.uci.edu/ml/datasets}.
Statlog (shuttle) has 9 features, 7 classes.
Coverype has 54 features, 7 classes.
      Adult has 89 features, 2 classes.
We use $T=5000$ samples for all datasets.
} %
We turn this into a bandit problem by defining
the actions to be the class labels,
and the reward is 1 if the correct label is predicted,
and is 0 otherwise.
Thus the cumulative reward is the number of correct classifications,
and the regret is the number of incorrect classifications.

\begin{figure*}
\centering
\includegraphics[height=2in]{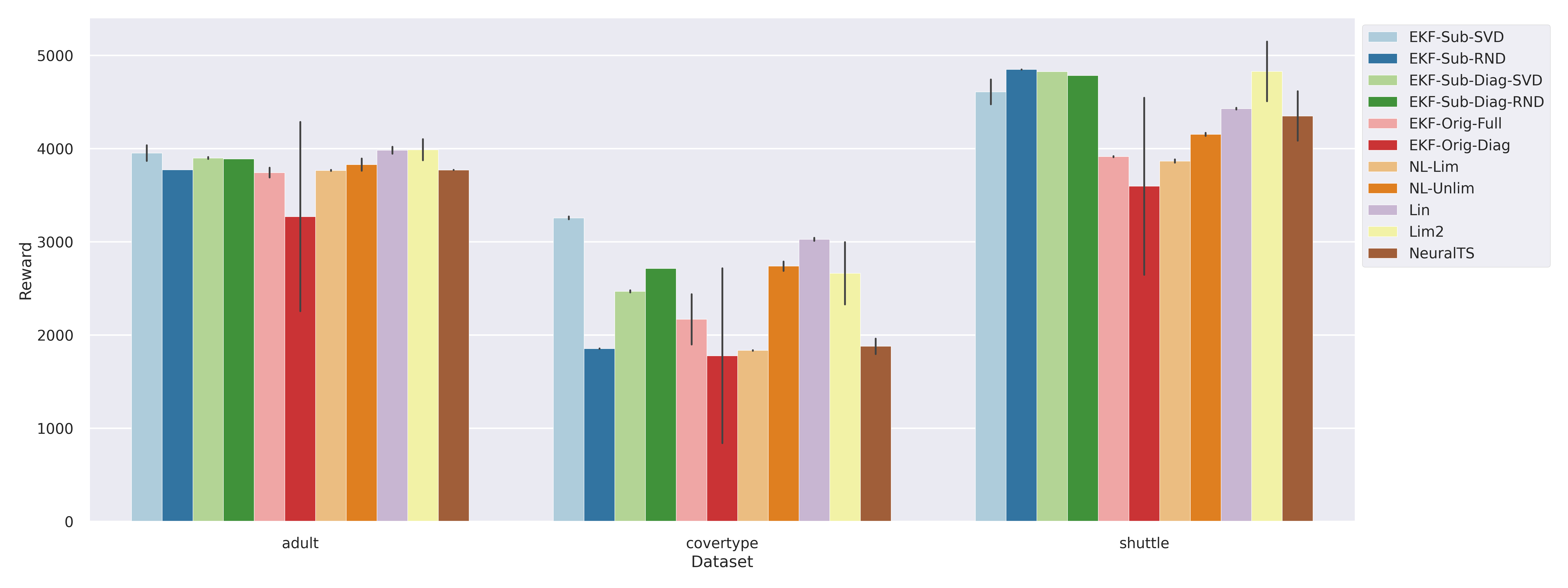}
\caption{
  Reward for various methods on 3 tabular datasets.
  The maximum possible reward for each dataset is 5000.
}
\label{fig:tabular-reward}
\end{figure*}

Following  prior work,
we use the multi-headed MLP in \cref{fig:mlp-multi-head},
with one hidden layer with  $\nhidden=50$ units
and \relu activations.
(The Neural-TS results are based on the multi-input model
in \cref{fig:mlp-repeat}.)
We use  $\npulls=20$ ``pulls'' per arm during the warmup phase
and run for $T=5000$ steps.
We run 10 random trials and report the mean reward,
together with the standard deviation.

We compare the following 11 methods:
EKF in a learned (SVD) subspace  (with full or diagonal covariance),
EKF in  a random subspace  (with full or diagonal covariance),
EKF in the original parameter space  (with full or diagonal covariance),
Linear,
Neural-Linear (with unlimited or limited memory),
LiM2,
and Neural-TS.
For the 6 EKF methods, we use our own code.\footnote{
Our code is available (in JAX) at
\url{https://github.com/probml/bandits}.
}
For LiM2 and Neural-TS,  we use the original code from the authors.\footnote{
LiM2 is available (in TF1) at
\url{
https://github.com/ofirnabati/Neural-Linear-Bandits-with-Likelihood-Matching}.
Neural-TS is available (in PyTorch)
at 
\url{https://github.com/ZeroWeight/NeuralTS}.
} %
For Linear and Neural-Linear methods, we reproduced the original code
from the authors in our own codebase.
All the hyperparameters are the same as in
the original papers/code
(namely \citep{Nabati2021} for Linear, Neural-Linear and Lim2,
and \citep{neuralTS} for Neural-TS).

We show the average reward for each method on each dataset
in \cref{fig:tabular-reward}.
(We  use $\nparamsSmall=200$ for all experiments,
which we found to work well.)
On the Adult dataset, all methods have similar performance,
showng that this is an easy problem.
On the Covertype dataset, we find that the best method
is EKF  in a learned (SVD) subspace with full covariance (light blue bar).
This is the only method to beat the linear baseline (purple).
On the Shuttle (Statlog) dataset,
we see that all the EKF subspace variants work well,
and match the accuracy of Lim2 while being much faster.
(We discuss speed in \cref{sec:time}.)
We see that EKF in the original parameter space peforms worse,
especially when we use a diagonal approximation (red).
We also see that limited memory version of neural linear (light orange)
is worse than unlimited memory (dark orange).

However, we also see that differences between most
methods are often rather small, and are often within the error bars.
We also noticed this with other examples from the Bandit Showdown benchmark
(results not shown).
We therefore believe this benchmark is too simple to be a reliable way of
measuring performance differences of neural bandit
algorithms (despite its popularity in the literature).
In  the sections below, we consider more challenging benchmarks,
where the relative performance differences are clearer.

\subsection{Recommender systems}
\label{sec:recsys}
\label{sec:movie}
\label{sec:movies}

\begin{figure}
\centering
\includegraphics[height=2in]{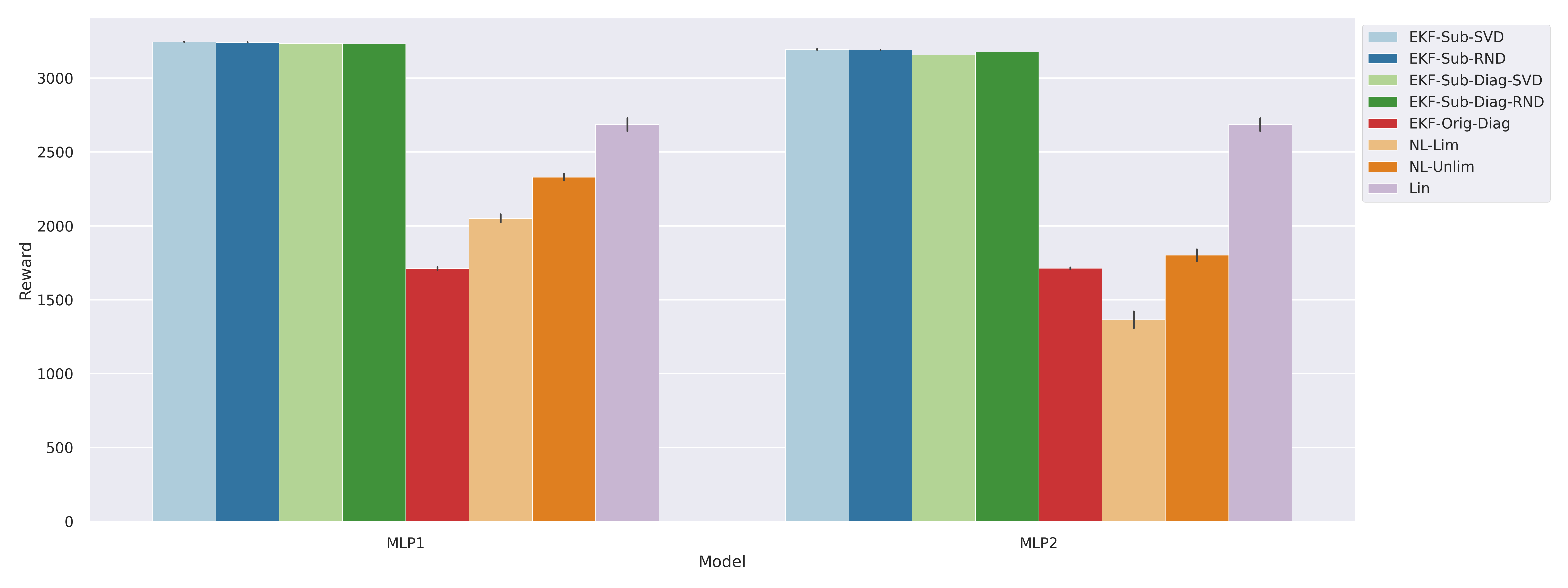}
\caption{
  Reward for various methods on the Movielens  dataset.
}
\label{fig:movielens-reward}
\end{figure}

One of the main applications of bandits is to recommender systems
(see e.g., \citep{Li10linucb,Guo2020bandits}).
Unfortunately, evaluating bandit policies in such systems
requires running a live experiment,
 unless we have a simulator or we use
off-policy evaluation methods such as those
in \citep{Li11}.
In this section, we build a simple simulator
by applying SVD to the MovieLens-100k dataset,
following the example in the TF-Agents library.\footnote{
See \url{https://blog.tensorflow.org/2021/07/using-tensorflow-agents-bandits-library-for-recommendations.html}.
}

In more detail,
we start with the MovieLens-100k dataset,
which has 100,000 ratings on a scale of 1--5 from 943 users on 1682 movies.
This defines a sparse $943 \times 1682$ ratings matrix,
where 0s correspond to missing entries.
We extract a subset of this matrix corresponding to the first 20 movies
to get a $943 \times 20$ matrix $\vX$.
We then compute the SVD of this matrix,
$\vX = \vU \vS \vV^\trans$,
and compute a dense low-rank approximation to it
$\hat{\vX} = \vU_K \vS_K \vV_K^\trans$.
(This is a standard approach to
matrix imputation, see e.g., \citep{Srebro03,Bell2007}).
We treat each user $i$ as a context,
represented by $\vu_i$,
and treat each movie $j$ as an action;
the reward for taking action $j$ in
in context $i$ is $X_{ij} \in \real$.
We follow the TF-Agents example
and use $K=20$, so the context has  20 features,
and there are also 20 actions (movies).

Having created this simulator, we can use it to evaluate various
bandit algorithms.
We use MLPs with 1 or 2 hidden layers, with 50 hidden units per layer.
Since the Lim2 and NeuralTS code
was not designed for this environment,
we restrict ourselves to the 9 methods
we have implemented ourselves.
We show the results in  \cref{fig:movielens-reward}.
On this dataset we see that the EKF subspace methods
perform the best (by a large margin), followed by linear,
and then neural-linear, and finally  EKF in the
original space (diagonal approximation).
We also see that the deeper model (MLP2) performs
worse than the shallower model (MLP1)
when using the neural linear approximation;
we attribute this to overfitting, due to not being Bayesian about
the parameters of the feature extractor.
By contrast, our fully Bayesian approach is robust to using
overparameterized models, even in the small sample setting.

\subsection{MNIST}
\label{sec:MNIST}

\begin{figure}
\centering
\includegraphics[height=2in]{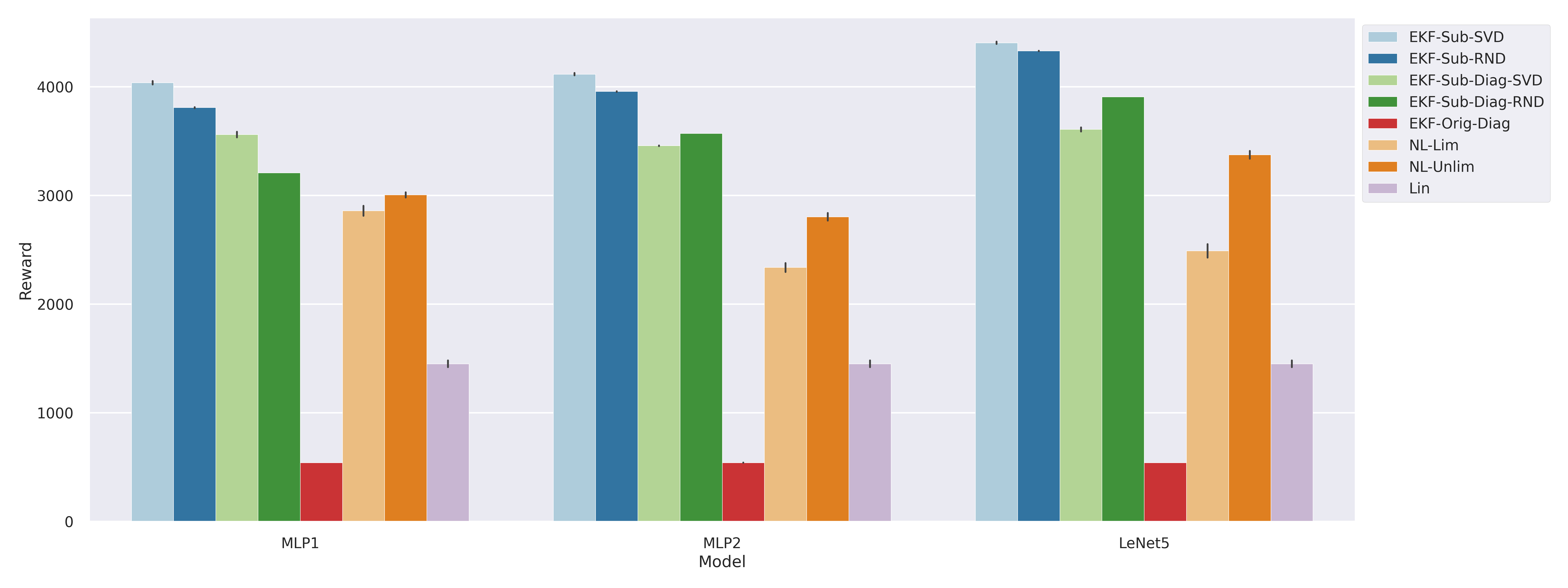}
\caption{
  Reward for various methods   on MNIST.
  The maximum possible reward is 5000.
}
\label{fig:mnist-reward}
\end{figure}

So far we have only considered low dimensional problems.
To check the scalability of our method, we applied it to MNIST,
which has 784 input features and 10 classes (actions).
In addition to a baseline
linear model,
 we consider three different kinds of deep neural network:
an MLP with 50 hidden units and 10 linear outputs (MLP1, with
$\nparams=39,760$ parameters),
an MLP with two layers of 200 hidden units each and 10 linear outputs (MLP2
with $\nparams=48,420$ parameters),
and a small convolutional neural network (CNN) known as LeNet5
\citep{LeCun98} with $\nparams=61,706$ parameters.

\eat{
MLP 50-A size = (39760,)
MLP 500-500-A size = (648010,)
MLP 200-200-A 48420
Lenet5 size = (61706,)
}

Not surprisingly, we find that the CNN works better than MLP2,
which works better than MLP1 (see \cref{fig:mnist-reward}).
Furthermore, for any given model,
we see that our EKF-subspace method
outperforms the widely used neural-linear method,
even though the latter has unlimited memory
(and therefore potentially takes $O(T^2)$ time).

For this experiment, we use a subspace
dimensionality of $\nparamsSmall=470$
(chosen using a validation set).
With this size of subspace, there is not a big difference
between using an SVD subspace and a random subspace.
However, using a full covariance in the subspace works better than a diagonal covariance
(compare blue bars with the green bars).
We see that all subspace methods work better than the neural linear baseline.
In the original parameter space, a full covariance is intractable,
and EKF with a diagonal approximation (red bar) works very poorly.

\subsection{Varying the subspace}
\label{sec:subspaceResults}

\begin{figure*}
\centering
\begin{subfigure}[b]{0.45\textwidth}
\centering
\includegraphics[height=1.75in]{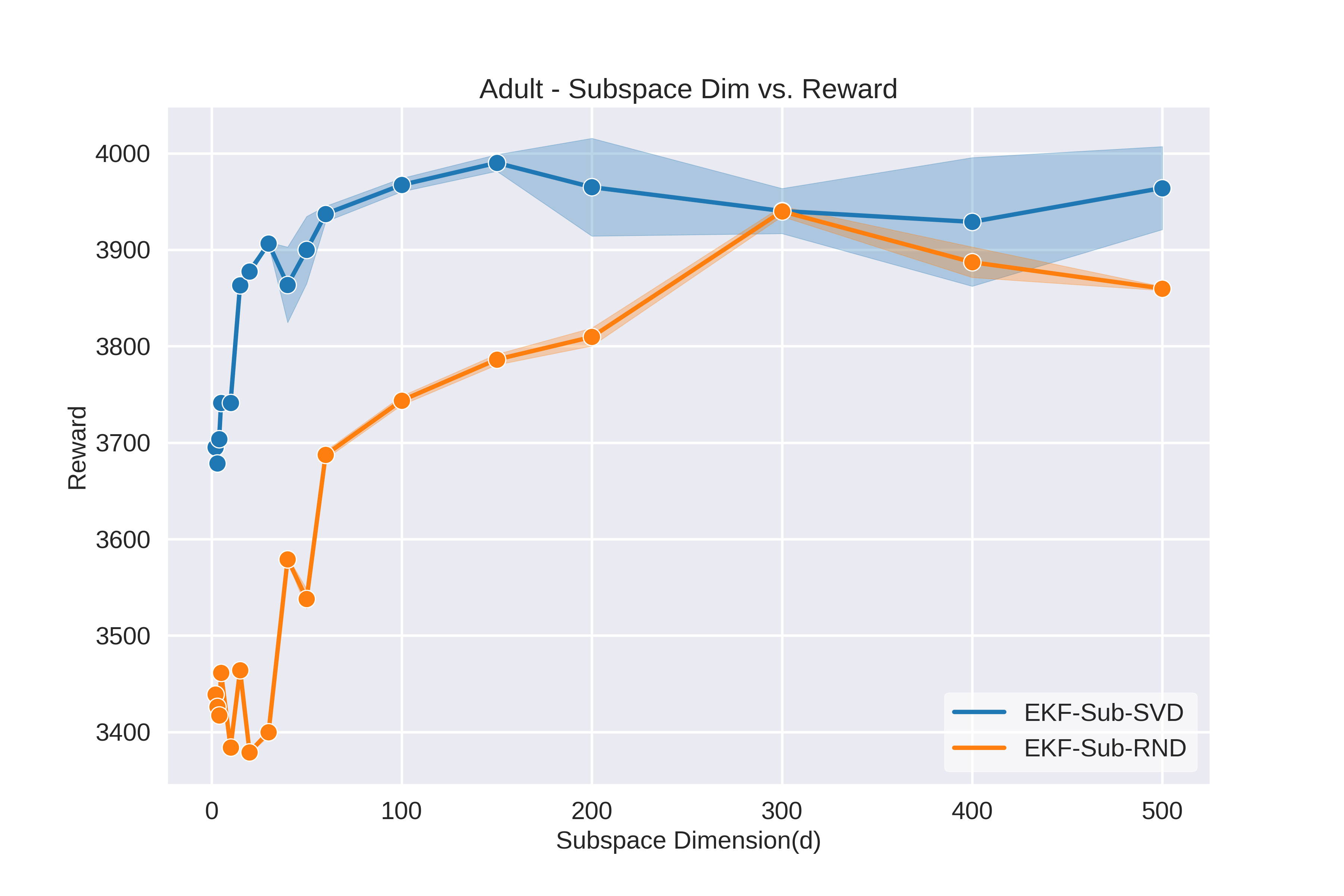}
\caption{ }
\end{subfigure}
~
\begin{subfigure}[b]{0.45\textwidth}
  \centering
  \includegraphics[height=1.75in]{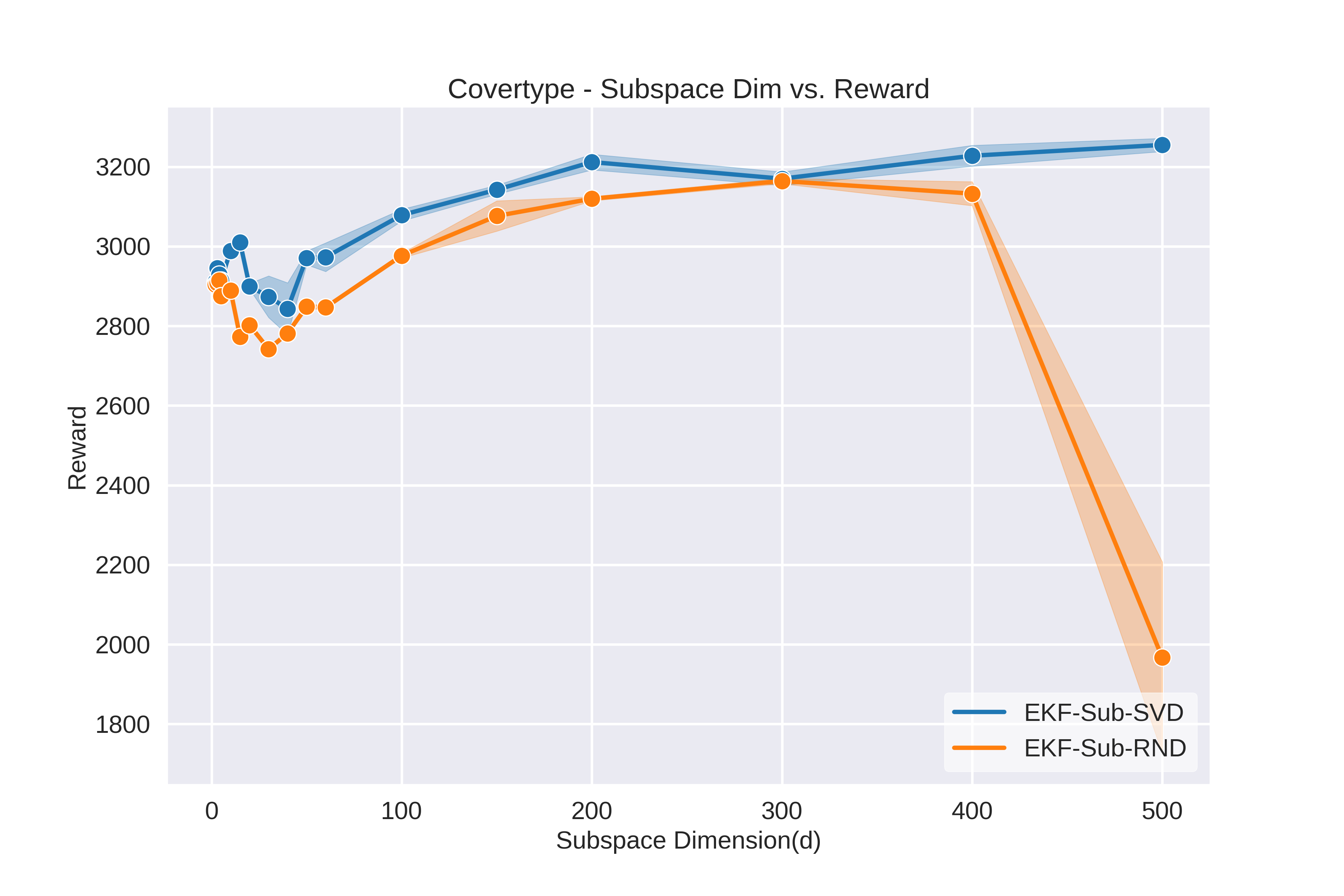}
\caption{ }
\end{subfigure}
\caption{
  Reward vs dimensionality of the subspace
  on (a) Adult, (b) Covertype.
  Blue estimates the subspace using SVD, orange uses a random subspace.
}
\label{fig:dim-reward}
\end{figure*}

\eat{
\begin{figure*}
\centering
\begin{subfigure}[b]{0.3\textwidth}
\centering
\includegraphics[height=1.2in]{arxiv-figures/adult_sub}
\caption{ }
\end{subfigure}
~
\begin{subfigure}[b]{0.3\textwidth}
  \centering
  \includegraphics[height=1.2in]{arxiv-figures/covertype_sub}
\caption{ }
\end{subfigure}
~
\begin{subfigure}[b]{0.3\textwidth}
  \centering
  \includegraphics[height=1.2in]{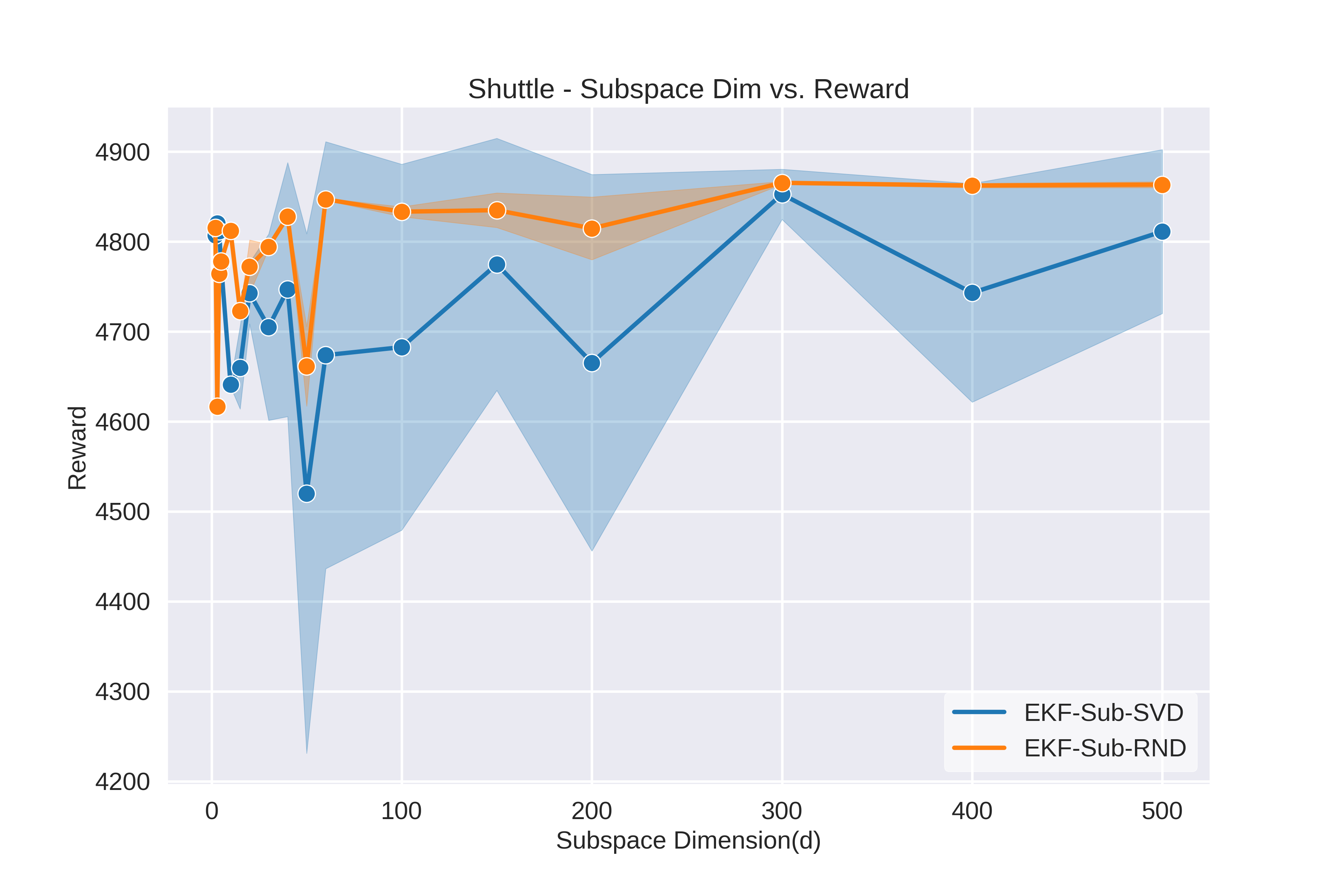}
\caption{ }
\end{subfigure}
\caption{
  Reward vs dimensionality of the subspace
  on (a) Adult, (b) Covertype, (c) Statlog.
  Blue estimates the subspace using SVD, orange uses a random subspace.
}
\label{fig:dim-reward}
\end{figure*}
}

A critical component of our approach is
how we estimating the parameter subspace
matrix $\vA \in \real^{D \times d}$.
As we explained in \cref{sec:subspace}, we have two different approaches
for computing this:
randomly or based on SVD applied
to the parameter iterates computing by gradient descent during the warmup phase.
We show the performance vs $d$ for these two approaches
in \cref{fig:dim-reward} for a one-layer MLP
with $D \sim 40k$ parameters
on some tabular datasets.
We see two main trends:
SVD is usually much better than random, especially in low dimensions;
and performance usually increases with $\nparamsSmall$, and then
either plateaus or even drops.
The drop in performance with increasing dimensionality is odd,
but is consistent with the results in
\citep{Larsen2021degrees}, who noticed exactly the same effect.
We leave investigating the causes of this to future work.

\subsection{Time and space complexity}
\label{sec:time}

\begin{figure*}
\centering
\includegraphics[height=2in]{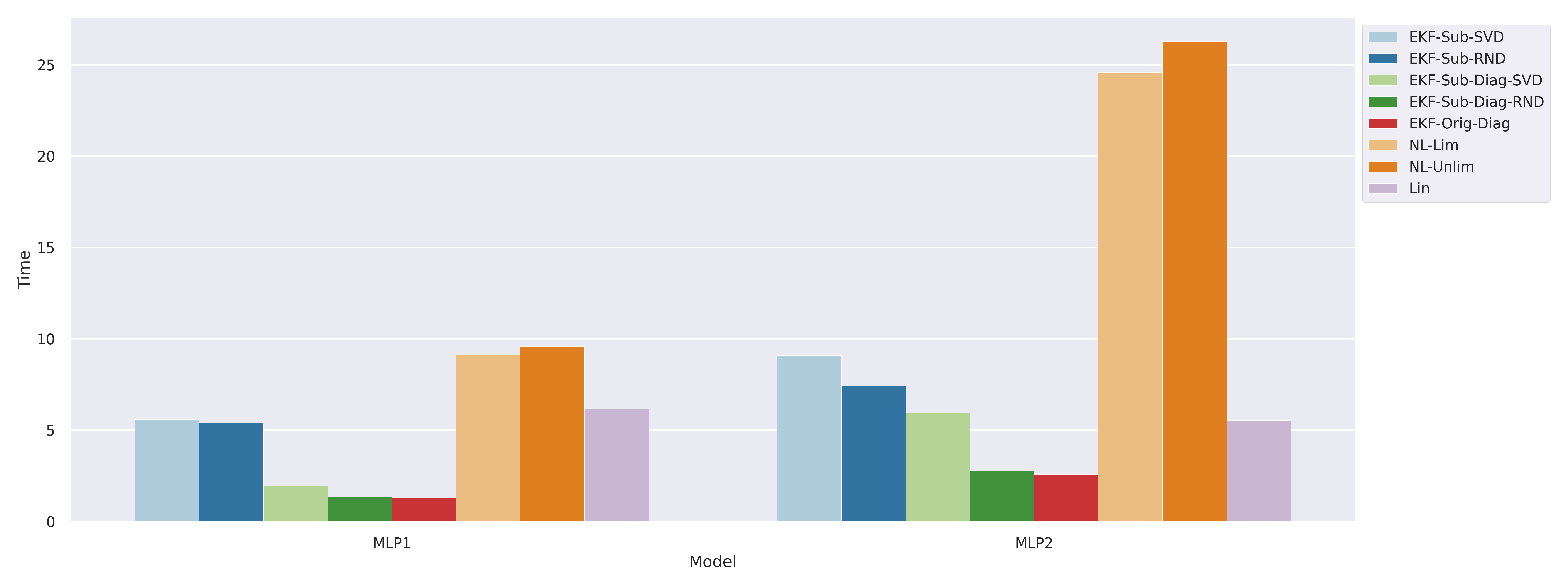}
\caption{
  Running time (CPU seconds) for 5000 steps using various methods
  on MovieLens.
}
\label{fig:timeMovies}
\end{figure*}

\begin{figure*}
\centering
\includegraphics[height=2in]{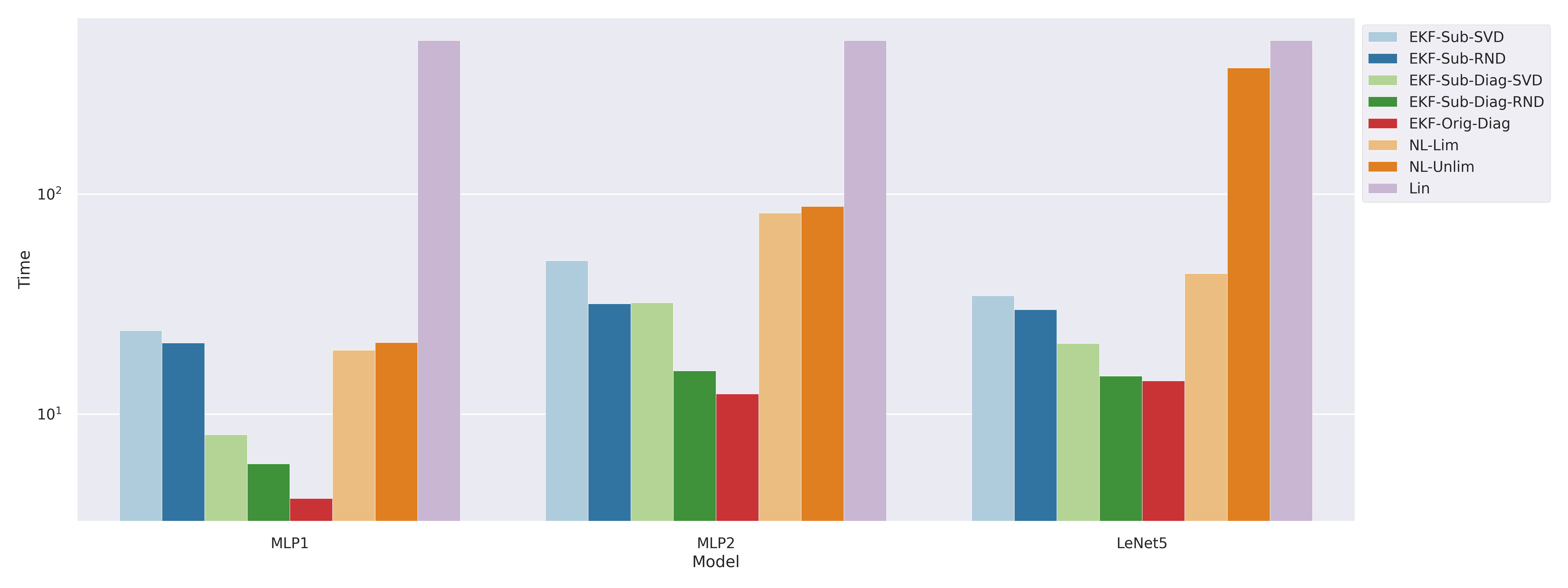}
\caption{
  Running time (CPU seconds) for 5000 steps using various methods
  on MNIST. Note the vertical axis is logarithmic.
}
\label{fig:timeMNIST}
\end{figure*}

\eat{
\begin{figure*}
\centering
\begin{subfigure}[b]{0.45\textwidth}
\centering
\includegraphics[height=1.5in]{\figdir/statlog-time-vertical3.pdf}
\caption{ }
\label{fig:tabular-time}
\end{subfigure}
~
\begin{subfigure}[b]{0.45\textwidth}
\centering
\includegraphics[height=1.2in]{\figdir/mnist_time_log_grouped.pdf}
\caption{ }
\label{fig:mnist-time}
\end{subfigure}
\caption{
  Running time (econds) for 5000 steps using various methods.
  (a)
  Statlog data with 1 layer MLP (except for linear).
  (b) MNIST data with different architectures.
  Note the vertical scale is logarithmic.
}
\label{fig:time}
\end{figure*}
}

One aspect of bandit algorithms that has been overlooked in the literature
is their time and space complexity, which is important in many practical applications,
like recommender systems or robotic systems,
that may run indefinitely (and hence need bounded memory)
and need a fast response time.
We give the asymptotic complexity of each method
in \cref{tab:methods}.
In \cref{fig:timeMovies}, we show the empirical
wall clock time  for each method when applied to the MovieLens dataset.
We see the following trends:
Neural-linear methods (orange) are the slowest,
with the limited memory version usually being slightly faster
than the unlimited memory version, as expected.
The EKF subspace methods 
are the second slowest, with SVD slightly slower  than RND,
and full covariance (blue) slower than diagonal (green).
Finally, the fastest method is diagonal EKF in the 
original parameter space;
however, the performance (expected reward) of this method is poor.
It is interesting to note that our subspace models are faster than the linear baseline;
this is because we only have to invert a $d \times d$ matrix,
instead of inverting $\nactions$
matrices, each of size $\nfeatures \times \nfeatures$.

In \cref{fig:timeMNIST}, we show the empirical
wall clock time for each method when applied to the MNIST dataset.
The relative performance trends (when viewed on a log scale)
are similar to the MovieLens case.
However, the linear baseline is much slower than most other methods,
since it works with the 784-dimensional input features,
whereas the neural methods work with lower dimensional latent features.
We also see that the  neural  linear method is quite slow,
especially when applied to CNNs,
and even more so in the unlimited memory  setting.
(We could not apply Lim2 to MNIST since the code
is designed for the tabular datasets in the showdown benchmark.)

In addition to time constraints, memory is also a concern
for long-running systems. Most online neural bandit methods store
the entire past history of observations,
to avoid catastrophic forgetting.
If we limit SGD updates of the feature extractor
to a window of the last  $\memory=100$ observations,
performance drops (see e.g., \cref{fig:tabular-reward}).
The Lim2 method attempts to solve this,
but is very slow, as we  have seen.
Our subspace EKF method is both fast and memory efficient.

\eat{
\subsection{Off-policy evaluation on the Yahoo! dataset}
\label{sec:ads}

TBD.

}

\section{Discussion}
\label{sec:discuss}
\label{sec:discussion}

We have shown that we can perform efficient online
Bayesian inference for large neural networks by applying
the extended Kalman filter to a low dimensional version
of the parameter space. 
In future work, we would like to apply the method
to other sequential decision problems,
such as Bayesian optimization and active learning.
We also intend to extend it 
to  Bernoulli and other GLM bandits \citep{Filippi2010}.
Fortunately, we can generalize the EKF (and hence our method)
to work with the exponential family,
as explained in \citep{Ollivier2018}. 

\eat{
Another extension we hope to pursue in the future
is to replace the EKF algorithm with particle filtering,
possibly with an EKF proposal;
this would be  similar to  \citep{deFreitas00},
but would perform inference in a subspace.
The non-parametric nature of PF could give improve posterior approximations,
which could in turn result in better decision making and lower regret
\citep{Phan2019}.
}

Finally, a note on societal impact.
Our method makes online Bayesian inference for neural networks more tractable,
which could increase their use. We view this as a positive thing, since
Bayesian methods can express uncertainty, and may be less prone to
making confident but wrong decisions \citep{Bhatt2021}.
However,  we acknowledge that bandit algorithms are often
used for recommender systems and online advertising, which can have some
unintended harmful societal effects \citep{Milano2020}.

\subsubsection*{Acknowledgements}

We would like to thank
Luca Rossini,
Alex Shestopaloff
and
Efi Kokiopoulou
for helpful comments on an earlier draft of the paper.

\newpage

\appendix

\section{More details on the methods}
\label{sec:app}

\subsection{Linear bandits}
\label{app:Linear}
\label{app:linearBandit}

In this section, we discuss how to do belief updating for a linear bandit,
where the reward model has the form
$\freward(\state,\action;\vtheta) = \vw_{\action}^\trans \state$,
where $\vtheta=\vW$ are the parameters.
(We ignore the bias term, which can be accomodated by augmenting
the input features $\state$ with a constant 1.)
To simplify the notation,
we give the derivation for a  single arm.
In practice, this procedure is repeated
separately for each arm,  using the contexts and rewards
for the time periods where that arm was used.

\subsubsection{Known variance $\sigma^2$}
\label{sec:knownSigma}

For now, we assume the observation noise $\sigma^2$ is known.
We start with
the uniformative prior $\belief_{0} = \gauss(\vw|\vmu_{0},\vSigma_{0})$,
where $\vmu_{0}=\vzero$ is the prior mean
and $\vSigma_{0}=(1/\epsilon) \vI$ is the prior covariance
for some small $\epsilon > 0$.
Let $\vX$ be the $N \times \nstates$ matrix of contexts
for this arm during the warmup period
(so $N=\npulls$ if we pull each arm $\npulls$ times),
and let $\vy$ be the corresponding $N \times 1$ vector of rewards.
We can compute the initial belief state based on the warmup data
by applying Bayes rule to the uninformative prior to get
\begin{align}
  p(\vw|\vX_{\nwarmup},\vy_{\nwarmup}) &= \gauss(\vw|\vmu_{\nwarmup},\vSigma_{\nwarmup}) \\
  \vSigma_{\nwarmup} &= (\vSigma_{0}^{-1} + \frac{1}{\sigma^2} \vX^\trans \vX)^{-1} \\
   \vmu_{\nwarmup}&=  \vSigma_{\nwarmup}(\vSigma_{0}^{-1} \vmu_{0} + \frac{1}{\sigma^2} \vX^\trans \vy)
\end{align}

After this initial batch update, we can perform incremental updates.
We can use the Sherman-Morrison formula for rank one updating 
to efficiently compute the new covariance, without any matrix inversions:
\be
\vSigma_t = (\vSigma_{t-1}^{-1} + \frac{1}{\sigma^2} \vx_t \vx_t^\trans)^{-1}
= \vSigma_{t-1} -  \frac{\vSigma_{t-1} \vx_t \vx_t^\trans \vSigma_{t-1}}
{\sigma^2 + \vx_t^\trans \vSigma_{t-1} \vx_t}
\ee
To compute the mean,  we will assume $\vmu_0=\vzero$
and $\vSigma_0 = \kappa^2 \vI$.
Then we have
\begin{align}
  \vmu_t &=
  \frac{1}{\sigma^2} \vSigma_t \vX^\trans \vy  =   \frac{1}{\sigma^2} \vSigma_t \vpsi_t \\
  \vpsi_t &= \vpsi_{t-1} + \vx_t y_t
\end{align}

An alternative (but equivalent) approach
is to use the recursive least squares (RLS) algorithm,
which is a special case of
the Kalman filter
(see e.g., \citep{Borodachev2016} for the derivation).
The updates are as follows:
\begin{align}
    e_t &= y_t - \vx_t^\trans \vmu_{t-1} \\
  s_t &= \vx_t^\trans \vSigma_{t-1} \vx_t + \sigma^2 \\
  \vk_t&= \frac{1}{s_t} \vSigma_{t-1} \vx_t \\
  \vmu_{t} &= \vmu_{t-1} + \vk_t e_t \\
  \vSigma_t &= 
    \vSigma_{t-1} - \vk_t \vk_t^\trans s_t
\end{align}
(Of course, we only update the belief state for the
arm that was actually pulled at time $t$.)

\subsubsection{Unknown variance $\sigma^2$}
\label{sec:unknownSigma}

Now we consider the case where $\sigma^2$ is also unknown,
as in \citep{Riquelme2018,Nabati2021}.
This lets the algorithm explictly represent uncertainty in the reward
for each action, which will increase the dynamic range of the sampled
parameters,
leading to more aggressive exploration.
We have noticed this gives improved results over fixing $\sigma$.

We will use a conjugate normal inverse Gamma prior
$\MVNIG(\vw,\sigma^2|\vmu_{0}, \vSigma_{0}, a_{0}, b_{0})$.
The batch update is as follows,
where $\vX$ is all the contexts for this arm up to $t$,
and $\vy$ is all the rewards for this arm up to $t$:
\begin{align}
p(\vw,\sigma^2|\vX, \vy)
&= \MVNIG(\vw,\sigma^2|\vmu_{t}, \vSigma_{t}, a_{t}, b_{t}) \\
\vSigma_{t} &= (\vSigma_{0}^{-1} + \vX^\trans \vX)^{-1} \\
\vmu_{t} &= \vSigma_{t}(\vSigma_{0}^{-1} \vmu_{t} + \vX^\trans \vy) \\
a_{t} &= a_{0} + \frac{N_t}{2} \\
b_{t} &= b_{0} + \frac{1}{2} \left(
\vy^\trans \vy + \vmu_{0}^\trans \vSigma_{0}^{-1} \vmu_{0}
- \vmu_{t} \vSigma_{t}^{-1} \vmu_{t} \right)
\end{align}
This matches Equations 1--2 of \citep{Riquelme2018}.\footnote{
There is a small typo in Equation 2 of \citep{Riquelme2018}:
 the $\vSigma_0$ should be inverted.
}
To sample from this posterior,
we first sample $\tilde{\sigma}^2 \sim \IG(a_{t},b_{t})$,
and then sample $\vw \sim \gauss(\vmu_{t}, \tilde{\sigma}^2 \vSigma_{t})$.

We can rewrite the above equations in incremental form as follows:
\begin{align}
p(\vw,\sigma^2|\data_{0:t})
&= \MVNIG(\vw,\sigma^2|\vmu_t, \vSigma_t,a_t, b_t) \\
\vSigma_t &= (\vSigma_{t-1}^{-1} + \vx_t \vx_t^\trans)^{-1} \\
\vmu_t &= \vSigma_t(\vSigma_{t-1}^{-1} \vmu_{t-1} + \vx_t y_t) \\
a_t &= a_{t-1} + \frac{1}{2} \\
b_t &= b_{t-1} + \frac{1}{2} \left(
y_t^2 + \vmu_{t-1}^\trans \vSigma_{t-1}^{-1} \vmu_{t-1}
- \vmu_t \vSigma_t^{-1} \vmu_t \right)
\end{align}
It is natural to want to derive a version of these equations
which avoids the matrix inversion at each step.
We can incrementally update $\vSigma_t$
without inverting $\vSigma_{t-1}$,
using Sherman-Morrison,
as in \cref{sec:knownSigma}.
However, computing $b_t$ needs access to $\vSigma_t^{-1}$.
Fortunately, we can generalize the Kalman filter
to the case where $V=\sigma^2$ is unknown,
as described in  \citep[Sec 4.6]{West97};
this avoids any matrix inversions.

To describe this algorithm,
let 
the likelihood at time $t$ be defined as follows:
\begin{align}
  p_t(y_t|\vw_t,V) &= \gauss(y_t|\vx_t^\trans \vw_t, V)
\end{align}
Let $\lambda=1/V$ be the observation precision.
To start the algorithm, we use the following prior:
\begin{align}
  p_0(\lambda) &= \Ga(\frac{\nu_0}{2}, \frac{\nu_0 \tau_0}{2}) \\
  p_0(\vw|\lambda) &= \gauss(\vmu_0, V \vSigma_0^*)
  \end{align}
where $\tau_0$ is the prior mean for $\sigma^2$,
and $\nu_0 > 0$ is the strength of this prior.
We now discuss the belief updating step.
We assume that the prior belief state at time $t-1$ is
\be
\gauss(\vw,\lambda|\data_{1:t-1})
= \gauss(\vz_w|\vmu_{t-1}, V \vSigma_{t-1}^*)
\Ga(\lambda|\frac{\nu_{t-1}}{2}, \frac{\nu_{t-1} \tau_{t-1}}{2})
\ee
The posterior is given by 
\be
\gauss(\vw,\lambda|\data_{1:t})
= \gauss(\vw|\vmu_{t}, V \vSigma_{t}^*)
\Ga(\lambda|\frac{\nu_{t}}{2}, \frac{\nu_{t} \tau_{t}}{2})
\ee
where
\begin{align}
  e_t &= y_t - \vx_t^\trans \vmu_{t-1} \\
  s_t^* &= \vx_t^\trans \vSigma_{t-1}^* \vx_t + 1\\
  \vk_t &= \frac{1}{s_t^*} \vSigma_{t-1}^* \vx_t \\
  \vmu_t &= \vmu_{t-1} + \vk_t e_t \\
  \vSigma^*_t &=   \vSigma_{t|t-1}^* - \vk_t \vk_t^\trans s_t^* \\
  \nu_{t} &= \nu_{t-1}  + 1 \\
  \nu_t \tau_t &= \nu_{t-1} \tau_{t-1} + e_t^2/s_t^*
  \end{align}
If we marginalize out $V$, the marginal distribution
for $\vz_t$ is a Student distribution.
However, for Thompson sampling,
it is simpler to sample $\tilde{\lambda} \sim \Ga(\frac{\nu_t}{2},\frac{\nu_t \tau_t}{2})$,
and then to sample $\vw \sim \gauss(\vmu_t, \tilde{\sigma}^2 \vSigma_t^*)$,
where $\tilde{\sigma}^2=1/\tilde{\lambda}$.

\subsection{Neural linear bandits}
\label{app:NeuralLinear}
\label{app:neuralLinear}

The neural linear model assumes
that $\freward(\state,\action;\params) = \vw_{\action}^\trans  \vphi(\state;\vV)$,
where $\vphi(\state;\vV)$ is the feature extractor.
It approximates
the posterior over all the parameters
by using a point estimate for $\vV$,
a Gaussian distribution for each $\vw_i$ (conditional on $\sigma^2_i$), and an inverse
Gamma distributon for each $\sigma^2_i$, i.e.,
\be
p(\vtheta|\data_{1:t}) =
\delta(\vV-\hat{\vV}_t)
\prod_{i=1}^{\nactions} \gauss(\vw_i|\vmu_{t,i}, \sigma^2_i \vSigma_{t,i})
\IG(\sigma^2_i|a_i,b_i)
\ee
where $\vtheta=(\vV,\vW,\va,\vb)$ are all the parameters,
and $\delta(\vu)$ is a delta function.
Furthermore, 
to avoid catastrophic forgetting,
we also need to store all of the previous observations,
so the belief state
has the form
$\belief_t = (\data_{1:t}, \hat{\vV}_t, \vmu_{t,1:\nactions}, \vSigma_{t,1:\nactions},
\va_{1:\nactions}, \vb_{1:\nactions})$.
The neural network parameters  are computed using SGD.
After updating $\hat{\vV}_t$,
we update the parameters of the Normal-Inverse-Gamma distribution
for the final layer weights $\vW$,
using the following equations
\begin{align}
  \vSigma_{i} &=  (\vSigma_{0,i}^{-1} + \vX_{i}^\trans \vX_i)^{-1}  \\
\vmu_{t,i} &=  \vSigma_{i}(\vSigma_{0,i}^{-1} \vmu_{0,i} + \vX_{i}^\trans \vy_i) \\
  a_{i} &=     a_{0,i} + \frac{N_{i}}{2} \\
  b_{i} &=    b_{0,i} + \half(\vy_i^\trans \vy_i + \vmu_{0,i}^\trans \vSigma_{0,i} \vmu_{0,i}
    - \vmu_{i}^\trans \vSigma_{i} \vmu_{i})    
\end{align}
where we define
$\vX_i = [\vphi_j: a_j=i]$ as the matrix whose
rows are the features $\vphi_j$ from time steps where action $i$ was taken,
and $\vy_i = [r_j: a_j=i]$ is the vector of rewards
from time steps where action $i$ was taken.
See \cref{algo:neuralLinear}
for the pseudocode.

\begin{algorithm}
\caption{Neural Linear.}
\label{algo:neuralLinear}
\dontprintsemicolon
\For{$t=(\nwarmup+1):T$}{
  $\state_t = \text{Environment.GetState}(t)$ \;
  $\tilde{\sigma}_i \sim \text{InverseGamma}(a_i, b_i)$ for all $i$ \;
  $\tilde{\vw}_i \sim \gauss(\vmu_i, \tilde{\sigma}_i \vSigma_i)$ for all $i$\;
  $a_t = \argmax_i \tilde{\vw}_i^\trans \vphi(\state_t;\vV_t)$ \;
   $\reward_t = \text{Environment.GetReward}(\state_t,\action_t)$ \;
 $\data_t = (\state_t, \action_t, \reward_t)$ \;
  \uIf{$t$ \rm{is an SGD update step}}{
    $\vtheta$ = \text{SGD}$(\vtheta, \data_{1:t})$ \;
    $\vV = \text{parameters-for-body}(\vtheta)$ \;
    Compute new features: $\vphi_j = \vphi(\state_j;\vV)$ for all $j \in \data_{1:t}$ \;
    \For{$i=1:\nactions$}{
    // Update sufficient statistics \; 
  $\vpsi_{i} = \sum_{j \leq t: a_t = i} \vphi_j y_j$ \;
  $\vPhi_{i} = \sum_{j \leq t: a_t=i} \vphi_j \vphi_j^\trans$  \;
   $R_{i}^2 =  \sum_{j \leq t: a_t=i}    y_t^2$ \;
    $N_{i} =    \sum_{j \leq t: a_t=i} 1$ \;
    // Update belief state \;
    $(\vmu_i, \vSigma_i, a_i, b_i) = \text{update-bel}(\vmu_{0,i}, \vSigma_{0,i}, a_{0,i}, b_{0,i},
    \vpsi_i, \vPhi_i, R_i^2, N_i)$
    }
  }
  \uElse{
    $i = a_t$ \;
  $\vpsi_{i} =    \vpsi_{i} + \vphi_t y_t$ \;
  $\vPhi_{i} =    \vPhi_{i} + \vphi_t \vphi_t^\trans$ \;
    $R_{i}^2 =     R_{i}^2 + y_t^2$ \;
    $N_{i} =     N_{i} + 1 $ \;
        $(\vmu_i, \vSigma_i, a_i, b_i) = \text{update-bel}(\vmu_{0,i}, \vSigma_{0,i}, a_{0,i}, b_{0,i},
    \vpsi_i, \vPhi_i, R_i^2, N_i)$
    }
}

\;
function update-bel($\vmu_{0,i}, \vSigma_{0,i}, a_{0,i}, b_{0,i}, \vpsi_i, \vPhi_i, R_i^2, N_i)$) \;
 $ \vSigma_{i} =  (\vSigma_{0,i}^{-1} + \vPhi_{i})^{-1}$  \;
$\vmu_{i} =  \vSigma_{i}(\vSigma_{0,i}^{-1} \vmu_{0,i} + \vpsi_i)$ \;
  $a_{i} =     a_{0,i} + \frac{N_{i}}{2}$ \;
  $b_{i} =    b_{0,i} + \half(R_{i}^2 + \vmu_{0,i}^\trans \vSigma_{0,i} \vmu_{0,i}
- \vmu_{i}^\trans \vSigma_{i} \vmu_{i}) $ \;
return $(\vmu_i, \vSigma_i, a_i, b_i)$
\end{algorithm}

\subsection{LiM2}
\label{app:LIM}
\label{app:Lim}
\label{app:LIM2}

In this section, we describe the LiM2
method of \citep{Nabati2021}.
It is similar to the neural linear method
except that the prior
($\vmu_{0,i}, \vSigma_{0,i})$ gets updated after each SGD step,
so as to not forget old information.
In addition, SGD is only applied to a rolling window
of the last $\memory$ most recent observations,
so the memory cost is bounded.
See \cref{algo:LIM} for the pseudocode.

\eat{
We denote this update by
\be
(\vtheta, \{\vmu_{0,i}, \vSigma_{0,i}\}) = \text{update-DNN-and-prior}(\vtheta,
\{\vmu_{0,i}, \vSigma_{0,i}\},
\data_{t-\memory:t})
\ee
See \cref{algo:LIM} for the pseudocode for this function.
We use this instead of
$\vtheta = \text{SGD}(\vtheta,\data_{1:t})$ in
\cref{algo:neuralLinear};
the rest of the code remains the same.
}

\begin{algorithm}
\caption{LiM2}
\label{algo:LIM}
\dontprintsemicolon
\For{$t=(\nwarmup+1):T$}{
  $\state_t = \text{Environment.GetState}(t)$ \;
  $\tilde{\sigma}_i \sim \IG(a_i, b_i)$ for all $i$ \;
  $\tilde{\vw}_i \sim \gauss(\vmu_i, \tilde{\sigma}_i \vSigma_i)$ for all $i$\;
  $a_t = \argmax_i \tilde{\vw}_i^\trans \vphi(\state_t;\vV_t)$ \;
   $\reward_t = \text{Environment.GetReward}(\state_t,\action_t)$ \;
  $\data_t = (\state_t, \action_t, \reward_t)$ \;
  $\calM_t = \text{push}(\data_t)$ \;
  \If{$|\calM_t| > \memory$}{$\calM_t = \text{pop}(\calM_t)$} 
  $(\vtheta, \{\vmu_{0,i}, \vSigma_{0,i}\})$ = \text{update-DNN-and-prior}
  $(\vtheta,\{\vmu_{0,i}, \vSigma_{0,i}\}, \calM_t)$ \;
  $\vV =\text{body}(\vtheta)$ \;
  Compute new features: $\vphi_j = \vphi(\state_j;\vV)$ for all $j \in \calM_t$ \;
    \For{$i=1:\nactions$}{
    // Update sufficient statistics \; 
  $\vpsi_{i} = \sum_{j \in \calM_t: a_t = i} \vphi_j y_j$ \;
  $\vPhi_{i} = \sum_{j \in \calM_t: a_t=i} \vphi_j \vphi_j^\trans$  \;
   $R_{i}^2 =  \sum_{j \in \calM_t: a_t=i}    y_t^2$ \;
    $N_{i} =    \sum_{j \in \calM_t: a_t=i} 1$ \;
    // Update belief state \;
    $(\vmu_i, \vSigma_i, a_i, b_i) = \text{update-bel}(\vmu_{0,i}, \vSigma_{0,i}, a_{0,i}, b_{0,i},
    \vpsi_i, \vPhi_i, R_i^2, N_i)$
    }
  }
\end{algorithm}

See \cref{algo:LIMupdate} for the pseudocode for the step
that updates the DNN and the prior on the last layer,
to avoid catastrophic forgetting.

\begin{algorithm}
\caption{LiM2 update step}
\label{algo:LIMupdate}
\dontprintsemicolon
Input: $\vtheta=(\vV,\vW)$, $\{\vmu_{0,i}, \vSigma_{0,i}\}$, $\data$ \;
\For{$P_1$ \rm{steps}}{
Sample mini batch $\data' = \{ \state_j, \action_j, \reward_j) : j =1:\batchsize\}$ from $\data$ \;
Compute old features: $\vphi_{j,\old} = \vphi(\state_j;\vV)$ for all $j \in \data'$ \;
$\vtheta$ = \text{SGD}($\vtheta$, $\data'$) \;
$\vV$ = \text{params-for-body}($\vtheta$), $\vW$ = \text{params-for-head}($\vtheta$) \;
Compute new features: $\vphi_j = \vphi(\state_j;\vV)$ for all $j \in \data'$ \;
\For{$i=1:\nactions$}{
  $\vSigma_{0,i}$ = \text{PGD}$( \vSigma_{0,i},
  \{\phi_{j,\old}: a_j=i\},  \{\phi_{j}: a_j=i\})$ \;
}
}
  $\vmu_{0,i} = \vw_i$ for each $i$ \;
Return $\vtheta$, $\{\vmu_{0,i}, \vSigma_{0,i}\}$ \;
\end{algorithm}

\begin{algorithm}
\caption{Projected Gradient Descent}
\label{algo:PGD}
\dontprintsemicolon
Input:  $\vA$, $\{\vphi_{j,old}\}$, $\{\vphi_{j}\}$ \\
$s_{j}^2 = \vphi_{j,\old}^\trans \vA \vphi_{j,\old}$ for all $j$ \;
$\vPhi_j = \vphi_j \vphi_j^\trans$ for all $j$ \;
\For{$P_2$ \rm{steps}}{
    $\vg = 2 \sum_{j} (\tr(\vA \vPhi_j) - s_j^2) \vPhi_j$ \\
  $\vA = \vA - \eta \vg$ \\
  $(\vLambda,\vV) = \text{eig}(\vA)$ \\
  $\calN = \{ k: \lambda_k < 0 \}$ \\
  $\vLambda[k,k] = 0 \text{ for all } k \in \calN$ \\
  $\vV[:,k] = 0 \text{ for all } k \in \calN$ \\
  $\vA = \vV \vLambda \vV^\trans$
}
Return $\vA$\;
\end{algorithm}

See \cref{algo:PGD}
for the projected gradient descent (PGD) step,
which solves 
a semi definite program to optimize the new covariance.

\subsection{Neural Thompson}
\label{app:NeuralTS}
\label{app:neuralTS}

In this section,
we discuss the ``Neural Thompson Sampling'' method of
\citep{neuralTS}.
We follow the presentation of
\citep{Levecque2021}, that shows the connection with linear
TS.

First consider the linear model
$r_{t,a} = \vx_{t,a}^\trans \vw$.
We assume $\sigma^2$ is fixed, $\vSigma_0 = \kappa^2 \vI$,
$\vmu_0 = \vzero$, and $\lambda = \frac{\sigma^2}{\kappa^2}$.
Recall from \cref{sec:knownSigma}
that  the posterior over the parameters
is given by
\begin{align}
  \vSigma_t &=
  \left[ \frac{1}{\sigma^2} ( \sigma^2 \vSigma_0^{-1}
    + \sum_{j=1}^t \vx_j \vx_j^\trans     \right]^{-1}
  =  \sigma^2 \left[
    \underbrace{\lambda \vI
      + \sum_{j=1}^t \vx_j \vx_j^\trans}_{\vB_t} \right]^{-1} \\
  \vmu_t &= \frac{1}{\sigma^2} \vSigma_t \vpsi_t = \vB_t^{-1} \vpsi_t
   = \vB_t^{-1} \sum_{j=1}^T \vx_j \reward_j
\end{align}
Thus the posterior over the parameters is given by
\be
p(\vw|\data_{1:t})
= \gauss(\vw|\vmu_t, \lambda \kappa^2 \vB_t^{-1})
\ee
The induced  posterior predictive
distribution over the reward is given by
\begin{align}
  p(\reward|\state,\action,\data_{1:t-1})
   &= \gauss(\reward|\mu_{t,a},v_{t,a}) \\
  \mu_{t,a}    &= \vx_{t,a}^\trans \expect{\vw} = \vx_{t,a}^\trans \vmu_{t-1} \\
  v_{t,a}    &= \vx_{t,a}^\trans \var{\vw} \vx_{t,a}
  = \kappa^2 \lambda \vx_{t,a}^\trans \vB_{t-1}^{-1} \vx_{t,a}
\end{align}

Now consider the NTK case.
We replace $\vx_{t,a} $ with
\be
\vphi_{t,a} = \frac{1}{\sqrt{\nhidden}}
 \nabla_{\params} \freward(\state,\action;\params)|_{\params_{t-1}}
 \ee
 which is the gradient
 of the neural net (an MLP with $\nhidden$ units per layer).
If we set $\kappa^2=1/\nhidden$, then the posterior predictive
distribution for the reward becomes
\begin{align}
  p(\reward|\state,\action,\data_{1:t})
   &= \gauss(\reward|\mu_{t,a},v_{t,a}) \\
  \mu_{t,a}    &= \freward(\state_t, a; \params_{t-1}) \\
  v_{t,a}    &= 
   \lambda \vphi_{t,a}^\trans \vB_{t-1}^{-1} \vphi_{t,a}
\end{align}
where
\begin{align}
\vB_t &= \vB_{t-1} +
\vphi(\state_t, \action_t; \vtheta_t) \vphi(\state_t, \action_t; \vtheta_t)^\trans
\end{align}
and we initialize with $\vB_0 = \lambda \vI$.
We sample the reward from this distribution for each action $a$,
and then the greedy action is chosen.

\subsection{EKF}
\label{app:EKF}

In this section, we describe the extended Kalman filter
(EKF) formulation in more detail.
Consider the following nonlinear Gaussian state space model:
\begin{align}
\vz_t &= \vf_{t}(\vz_{t-1}) +  \gauss(\vzero,\vQ_{t}) \\
\vy_t &= \vh_t(\vz_{t}) + \gauss(\vzero,\vR_t)
\end{align}
where $\hmmhid_t \in \real^{\NlatentKF}$ is the hidden state,
$\hmmobs_t \in \real^{\NobsKF}$ is the observation,
$\vf_t: \real^{\NlatentKF} \ra \real^{\NlatentKF}$ is the dynamics model,
and
$\vh_t: \real^{\NlatentKF} \ra \real^{\NobsKF}$ is the observation model.
The EKF linearizes the model at each step
by computing the following Jacobian matrices:
\begin{align}
  \vF_{t} &= \frac{\partial \vf_t(\vz)}{\partial \vz}|_{\vmu_{t-1}} \\
  \vH_{t} &= \frac{\partial \vh_t(\vz)}{\partial \vz}|_{\vmu_{t|t-1}} 
\end{align}
(These terms are  easy to compute using standard libraries such as JAX.)
The updates then become
\begin{align}
  \vmu_{t|t-1} &= \vf(\vmu_{t-1}) \\
  \vSigma_{t|t-1}&=  \vF_{t} \vSigma_{t-1} \vF_{t} + \vQ_{t} \\
  \ve_t &= \hmmobs_t - \vh(\vmu_{t|t-1}) \\
  \vS_t &= \vH_t \vSigma_{t|t-1}   \vH_t^\trans + \vR_t \\
  \vK_t &= \vSigma_{t|t-1} \vH_t \vS_t^{-1} \\
  \vmu_t &= \vmu_{t|t-1} + \vK_t \ve_t \\
  \vSigma_t &=    \vSigma_{t|t-1} - \vK_t \vH_t \vSigma_{t|t-1}
=  \vSigma_{t|t-1} - \vK_t \vS_t \vK_t^\trans
  \end{align}
(In the case of Bernoulli bandits,
we can use the exponential family formulation of the EKF
discussed in \citep{Ollivier2018}.)

The cost of the EKF is $O(\NobsKF \NlatentKF^2)$,
which can be  prohibitive for large state spaces.
In such cases, a natural approximation is to use a block diagonal
approximation.
Let us define the following
Jacobian matrices for block $i$:
\begin{align}
  \vF_{t}^i &= \frac{\partial \vf^i_t(\vz)}{\partial \vz}|_{\vmu_{t-1}} \\
  \vH_{t}^i &= \frac{\partial \vh^i_t(\vz)}{\partial \vz}|_{\vmu_{t|t-1}} 
\end{align}
We then compute the following  updates for each block:
\begin{align}
  \vmu_{t|t-1}^i &= \vf^i(\vmu_{t-1}) \\
  \vSigma_{t|t-1}^i &= (\vF_{t-1}^i)^\trans \vSigma_{t-1}^i \vF_t^i + \vQ_{t-1}^i \\
  \vS_t &=  \sum_i (\vH_t^i)^\trans \vSigma_{t|t-1}^i \vH_t^i + \vR_t \\
  \vK_t^i &= \vSigma_{t|t-1}^i \vH_t^i \vS_t^{-1} \\
  \vmu_t^i &= \vmu_{t|t-1} + \vK_t^i \ve_t \\
  \vSigma_t^i &=  \vSigma_{t|t-1}^i -\vK_t^i \vH_t^i \vSigma_{t|t-1}^i
\end{align}

Now we specialize the above equations to the setting
of this paper, where the latent state
is $\vz_t=\vtheta_t$, and the dynamics model $f_t$ is the identify function.
Thus the state space model becomes
\begin{align}
  p(\vtheta_t|\vtheta_{t-1}) &= \gauss(\vtheta_t|\vtheta_{t-1}, \vQ_t) \\
  p(\reward_t|\vx_t, \vtheta_{t}) &= \gauss(\reward_t|\freward(\vx_t,\vtheta_{t}), \vR_t) 
\end{align}
where $\vx_t=(\state_t,\action_t)$.
We set $\vR_t = \sigma^2$,
and $\vQ_t=\epsilon \vI$, to allow for a small amount
of parameter drift.
The EKF updates become
\begin{align}
  \vSigma_{t|t-1} &= \vSigma_{t-1} + \vQ_t \\
    \vS_t &=  \vH_t^\trans \vSigma_{t|t-1} \vH_t + \vR_t \\
  \vK_t &= \vSigma_{t|t-1} \vH_t \vS_t^{-1} \\
  \vmu_t &= \vmu_{t-1} + \vK_t \ve_t \\
  \vSigma_t &= \vSigma_{t|t-1} - \vK_t \vH_t \vSigma_{t|t-1}
\end{align}
The block diagonal version becomes
\begin{align}
  \vSigma_{t|t-1}^i &=  \vSigma_{t-1}^i  + \vQ_{t-1}^i \\
  \vS_t &=  \sum_i (\vH_t^i)^\trans \vSigma_{t|t-1}^i \vH_t^i + \vR_t \\
  \vK_t^i &= \vSigma_{t|t-1}^i \vH_t^i \vS_t^{-1} \\
  \vmu_t^i &= \vmu_{t-1}^i + \vK_t^i \ve_t \\
  \vSigma_t^i &=  \vSigma_{t|t-1}^i -\vK_t^i \vH_t^i \vSigma_{t|t-1}^i
\end{align}
This is called the ``decoupled EKF''
\citep{Puskorius1991,Puskorius2003}.

To match the notation in \citep{Puskorius2003},
let us define
$\vP_t=\vSigma_{t|t-1}$,
$\vw_t = \vmu_{t|t-1}$,
$\vA_t=\vS_t^{-1}$,
$\hat{\vH}_t^\trans = \vH_t$.
(Note that $\vA_t$ is a $N_o \times N_o$ matrix, so is a scalar
if $y_t \in \real$.)
\eat{
Then we can rewrite the above as follows:
\begin{align}
  \vA_t &= \left( \vR_t + \vH_t^\trans \vP_t \vH_t\right)^{-1} \\
  \vK_t &= \vP_t \vH_t \vA_t \\
  \vw_{t+1} &= \vw_t + \vK_t (\vy_t - \hat{\vy}_t) \\
  \vP_{t+1} &= \vP_t - \vK_t \hat{\vH}_t^\trans \vP_t + \vQ_t
\end{align}
}
Then we can rewrite the above as follows:
\begin{align}
  \vA_t &= \left( \vR_t + \sum_i (\vH_t^i)^\trans \vP_t^i \vH_t^i\right)^{-1} \\
  \vK_t^i &= \vP_t^i \vH_t^i \vA_t^i \\
  \vw_{t+1}^i &= \vw_t^i + \vK_t^i \ve_t\\
  \vP_{t+1}^i &= \vP_t^i - \vK_t^i (\hat{\vH}_t^i)^\trans \vP_t^i + \vQ_t^i
\end{align}


\printbibliography

\end{document}